\documentclass[justified, nobib]{tufte-handout}

\title{\mbox{Evaluating the ``Learning on Graphs'' Conference Experience}}
\author{Bastian Rieck and Corinna Coupette}
\date{}

\usepackage[english]{babel}

\usepackage[maxbibnames = 9]{biblatex}
\usepackage[super]{nth}
\usepackage{paralist}

\usepackage[defaultlines = 3, all]{nowidow}

\usepackage{graphicx}
\graphicspath{{figures}}
\usepackage{xspace}

\newcommand{\LoG}{LoG\xspace}
\newcommand{\question}[1]{\vspace*{1em}\noindent\textbf{#1}}

\begin{document}
	\maketitle
	
	\begin{abstract}
    With machine learning conferences growing ever larger, and
    reviewing processes becoming increasingly elaborate, more data-driven
    insights into their workings are required. In this report, we present the results of
    a survey accompanying the first ``Learning on Graphs''~(\LoG) Conference.
    The survey was directed to evaluate the submission and review process
    from different perspectives, including authors, reviewers, and area
    chairs alike.
	\end{abstract}
	
	\section{Motivation}

  The first ``Learning on Graphs''~(\LoG) Conference~(9--12 December,
  2022) was remarkable in more ways than one: starting from scratch,
  the conference aims to be \emph{the} place for graph learning
  research, making use of an advisory committee that consists of
  international experts in the field. Moreover, at is core, \LoG wants
  to be known for its exceptional review quality. With reviewing being
  an often-criticized process, marred by strong opinions that are held
  with high confidence, \LoG implemented three measures for improving
  review quality:
  \begin{inparaenum}[(i)]
    \item using sponsors to provide high monetary rewards for the best
      reviewers,
    \item vetting reviewers in advance, and
    \item assigning a smaller number of papers to the reviewers than other machine learning conferences.
  \end{inparaenum}
  The effectiveness of these measures can only be assessed holistically, which is why the authors of this report decided
  early on that a large-scale survey should accompany the conference.
  Such surveys are done regularly by conferences, but few, if any,
  appear to result in \emph{actionable changes} to the way conferences
  are run.

  Against this background, the results described in this report are aimed to engage the
  community, make the reviewing process more transparent, and, overall,
  serve as a way to challenge parts of the \emph{status quo} of running
  a conference. As our communities grow, our processes, too, must adapt. We cannot run the conferences of the \nth{21} century
  following procedures developed for community sizes of the \nth{20} century.

  \section{Related Work}

  Previous conferences, such as NeurIPS~2021, already rolled out surveys
  to assess certain aspect of the reviewing
  process~\autocite{Beygelzimer21a}, referencing a famous experiment at
  NeurIPS~2014~\autocite{Cortes21a}.
  Such surveys and experiments serve to highlight inconsistencies
  in the decision-making process \emph{per se}, and provide some encouragement
  to authors.\footnote{%
    To quote \textcite{Beygelzimer21a}: ``Finally, we would encourage
    authors to avoid excessive discouragement from rejections as there
    is a real possibility that the result says more about the review
    process than the paper.''
  }
  However, the size of NeurIPS and other conferences poses an
  obstacle to implementing large-scale changes, primarily because
  the program committee changes every year and knowledge transfer is not
  guaranteed. \LoG, by contrast, is positioned favorably because its
  research field is just emerging, being at least an order of magnitude
  smaller than NeurIPS. Moreover, the advisory committee guarantees
  a certain level of consistency in decisions. We hope that the results
  of our survey encourage other conferences to take a critical look at
  their underlying processes. To quote Lord Kelvin~\autocite[pp.\
  73--74]{Kelvin89}:
  \begin{quote}
  I often say that when you can measure what you are speaking about, and
  express it in numbers, you know something about it; but when you
  cannot measure it, when you cannot express it in numbers, your
  knowledge is of a meagre and unsatisfactory kind: it may be the
  beginning of knowledge, but you have scarcely, in your thoughts,
  advanced to the stage of science, whatever the matter may be.
  \end{quote}
  We hope that this survey begets knowledge that we may harness to improve
  future versions of \LoG and, perchance, other venues as well.

	\section{Results}
	
	To understand how participants experienced the \LoG conference, we distributed a survey of mostly closed, Likert-scaled questions to all authors, reviewers, and area chairs registered via OpenReview between from late November 2022 to mid February 2023.
	In this section, we present the results of the survey.\footnote{%
		For reproducibility, we make the response data and the code generating our analyses, excluding all sensitive information, available at the following DOI:  \href{https://doi.org/10.5281/zenodo.7875377}{10.5281/zenodo.7875377}.}
	In particular, for each part of the survey, 
	we visualize the results for each question, 
	providing the absolute numbers in the main visualization as well as the percentages in the marginals (for single-choice questions), 
	and using $n$ to indicate the number of respondents 
	(which might differ from the number of responses when multiple simultaneous responses were allowed).
	
	\subsection{Sample Composition}

  The survey was distributed to $162$ active submissions and we
  received $n = 183$ answers. Our breakdown of roles\footnote{
    To retain anonymity, our survey is not linked to paper IDs. We
    permit \emph{all} authors of a paper to respond to the survey.
  }
  indicates that~$92$ out of~$876$ authors responded~($10.5$\% of all authors or
  up to $50.2$\% of all active submissions), $118$ out of
  $372$ reviewers~($31.7$\%), and, finally, $3$ out of $46$ area
  chairs~($6.5$\%).

	
	\question{What roles did you have in the conference?}
	
	\includegraphics[width=\linewidth,keepaspectratio]{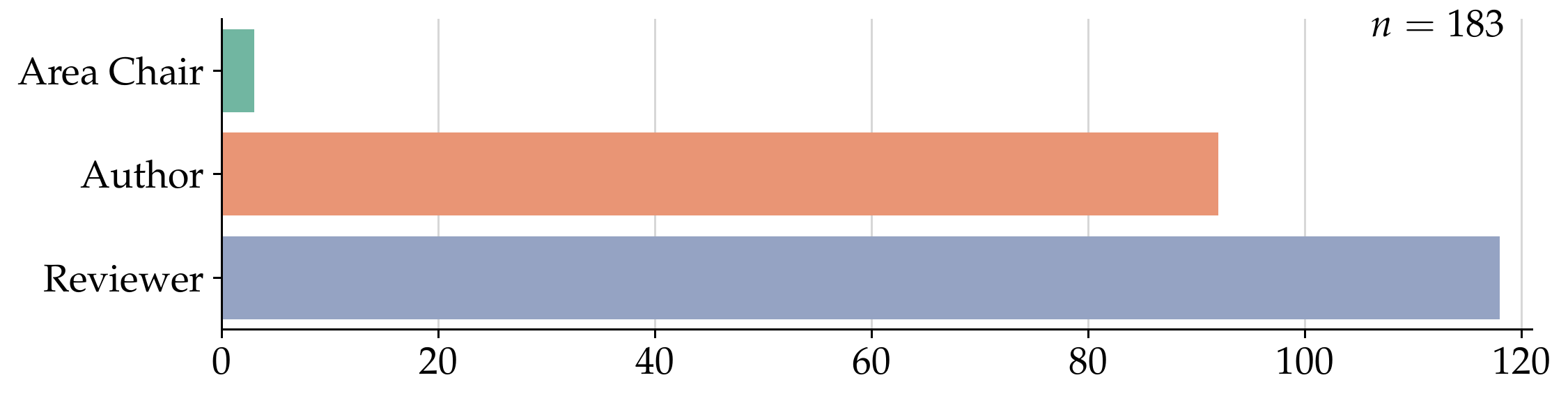}
	
	\includegraphics[width=\linewidth,keepaspectratio]{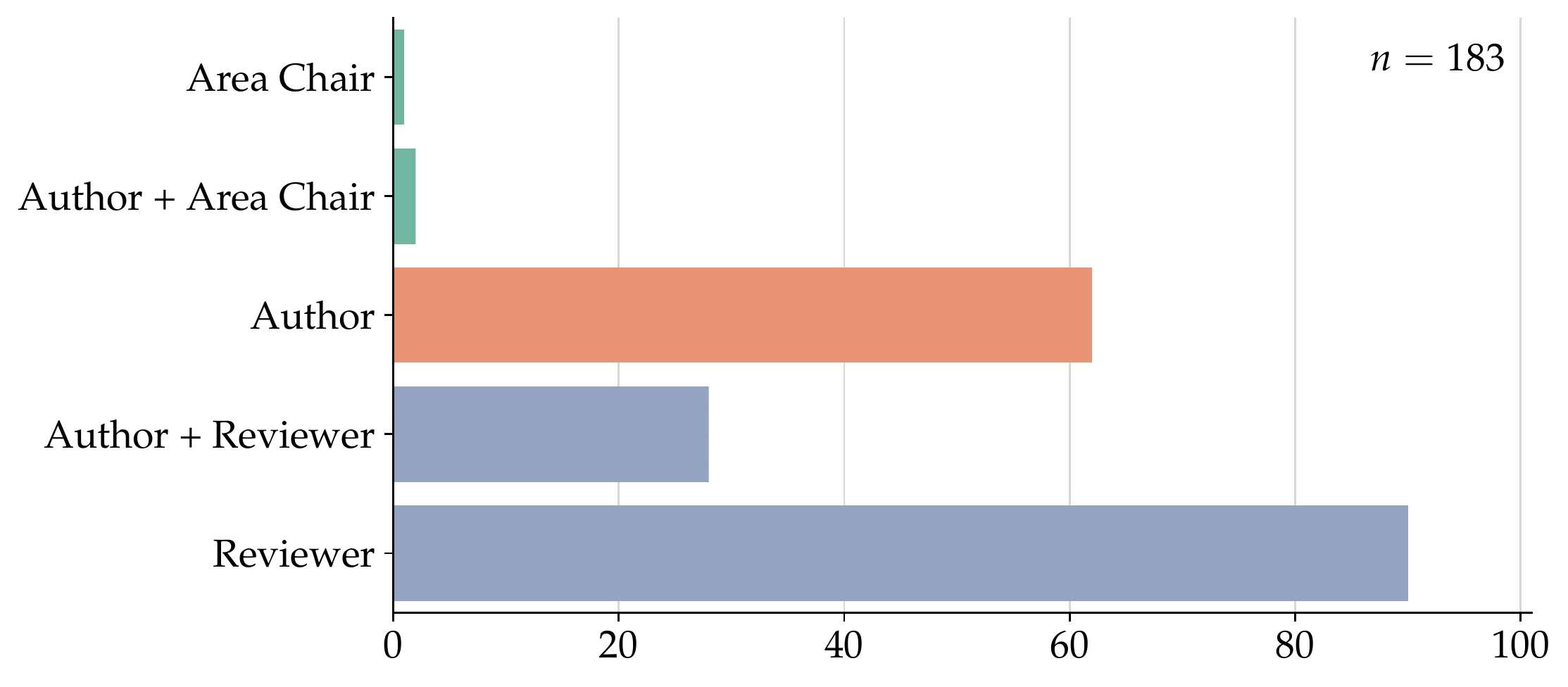}
		
	\subsection{Questions to Authors}

  We find that the overwhelming majority of authors is satisfied~(either
  moderately or extremely so) with the conference as well as the reviews
  that they received. When it comes to the experience of the rebuttal
  phase, authors tend to be slightly more neutral, but still positive
  overall. Interestingly, most authors rate the \emph{standards} of
  reviewers to be at least as high as those of comparable machine
  learning conferences---given that this was the first edition of the
  conference, this is an excellent outcome that vindicates the vetting
  process of reviewers. The fact that authors found the conference
  experience to be similar or better than comparable conferences is also
  an important signal that we consider to bode well for future editions.
	
	\question{As an author, how satisfied are you with the \emph{content} and \emph{quality} of the reviews?}	
	
		\marginnote[12pt]{%
		All respondents:\\
		\noindent\includegraphics[width=\linewidth]{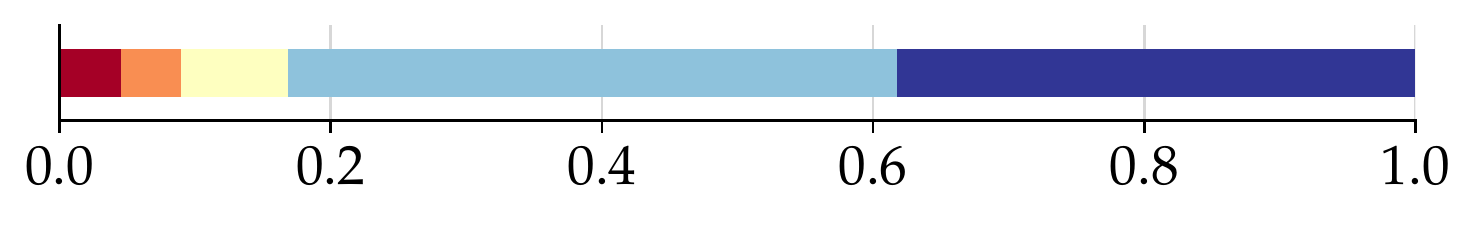}
		
		\noindent Author-only respondents:\\
		\noindent\includegraphics[width=\linewidth]{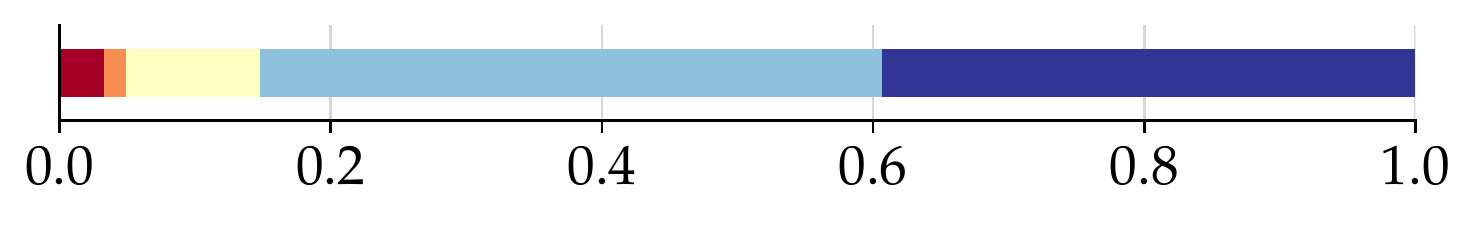}
		
		\noindent Author + \{reviewer, chair\} respondents:\\
		\noindent\includegraphics[width=\linewidth]{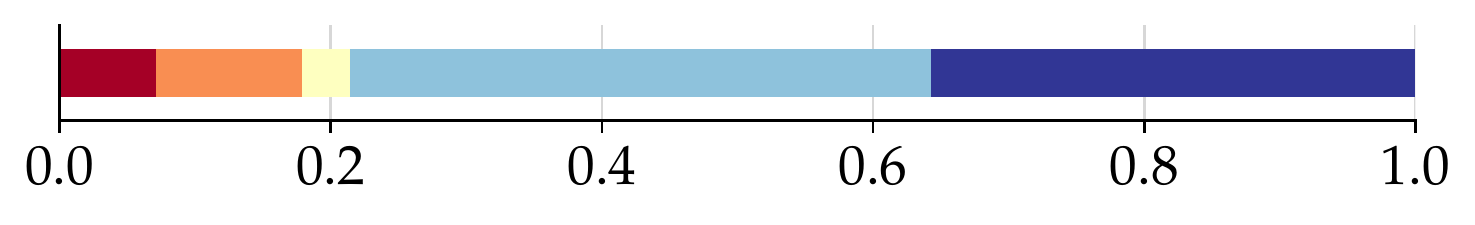}}
	
	\includegraphics[width=\linewidth,keepaspectratio]{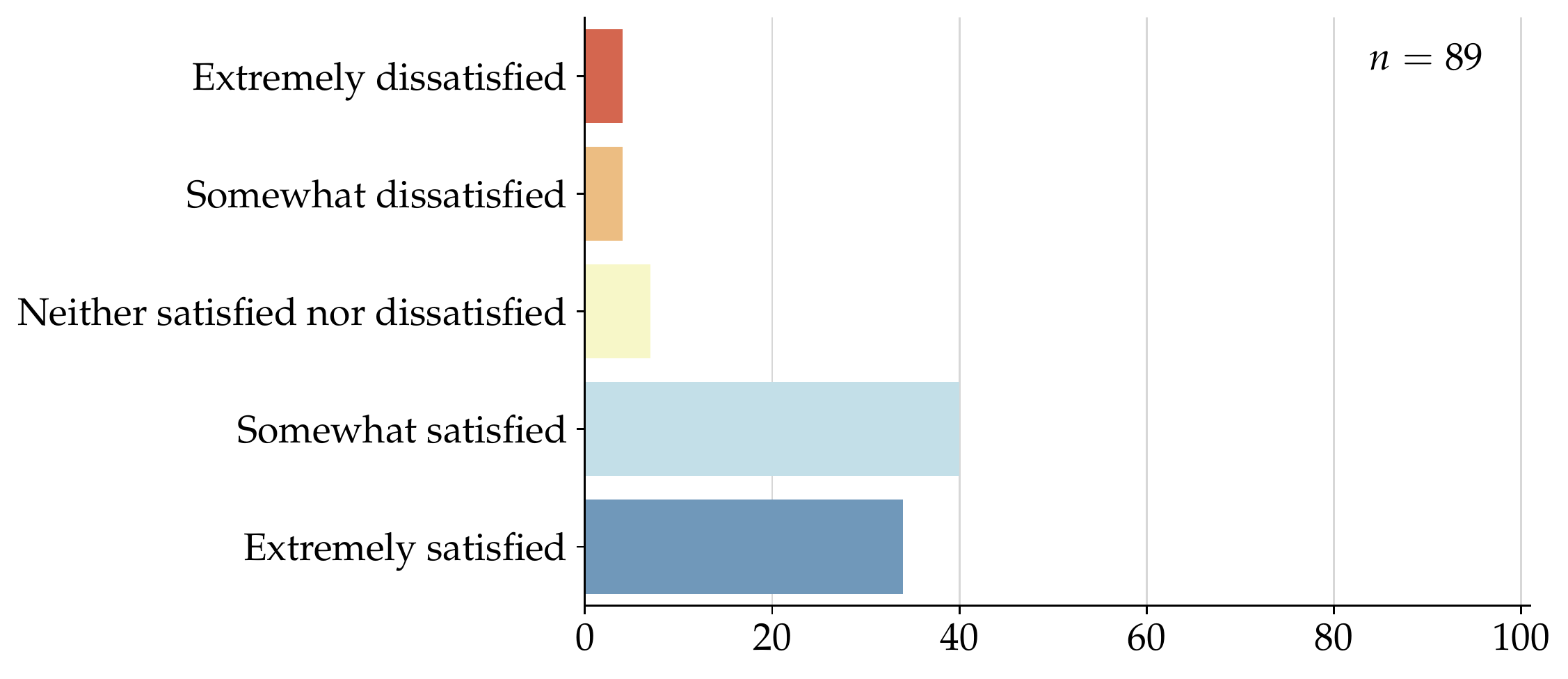}
	
	\clearpage
	
	\question{As an author, how satisfied are you with the \emph{tone} and \emph{style} of the reviews?}
	
		\marginnote[12pt]{%
		All respondents:\\
		\noindent\includegraphics[width=\linewidth]{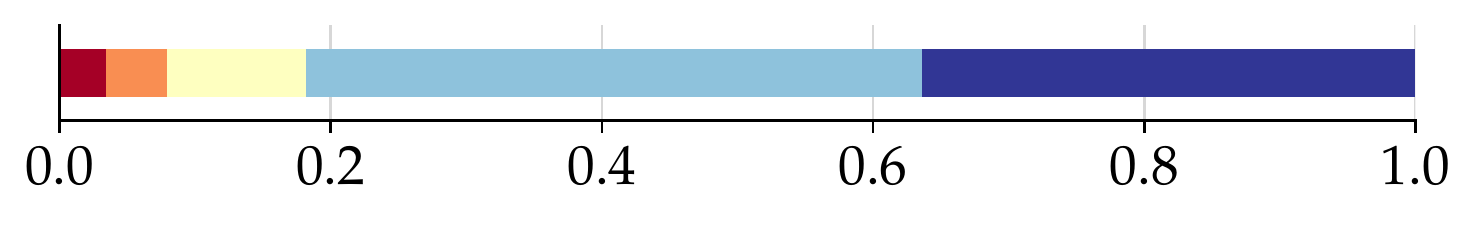}
		
		\noindent Author-only respondents:\\
		\noindent\includegraphics[width=\linewidth]{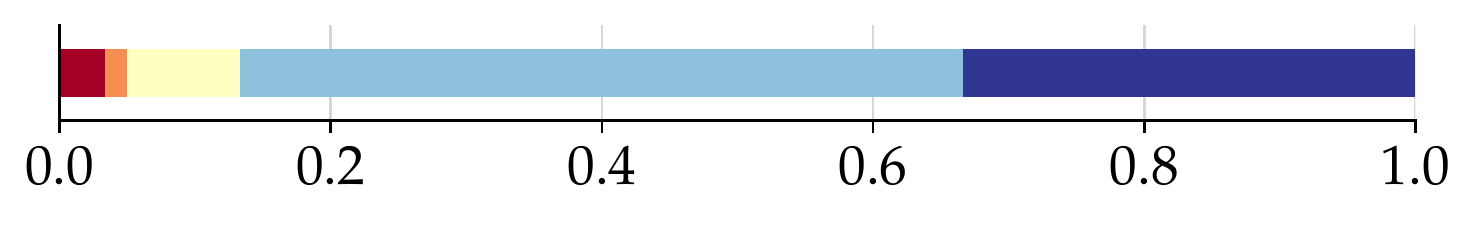}
		
		\noindent Author + \{reviewer, chair\} respondents:\\
		\noindent\includegraphics[width=\linewidth]{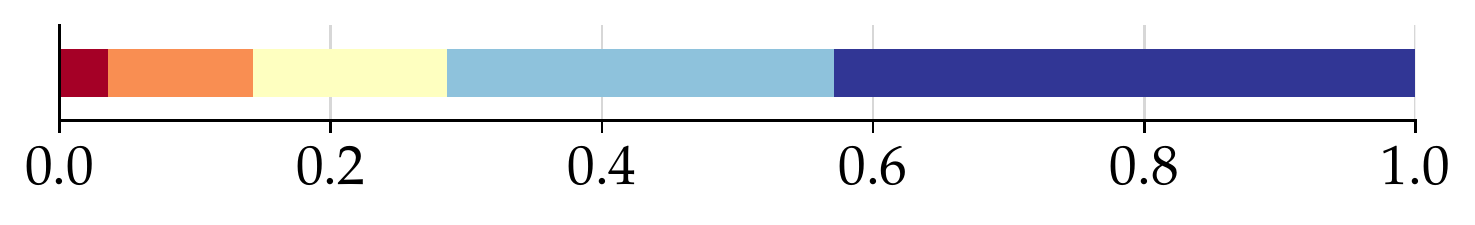}}
	
	\includegraphics[width=\linewidth,keepaspectratio]{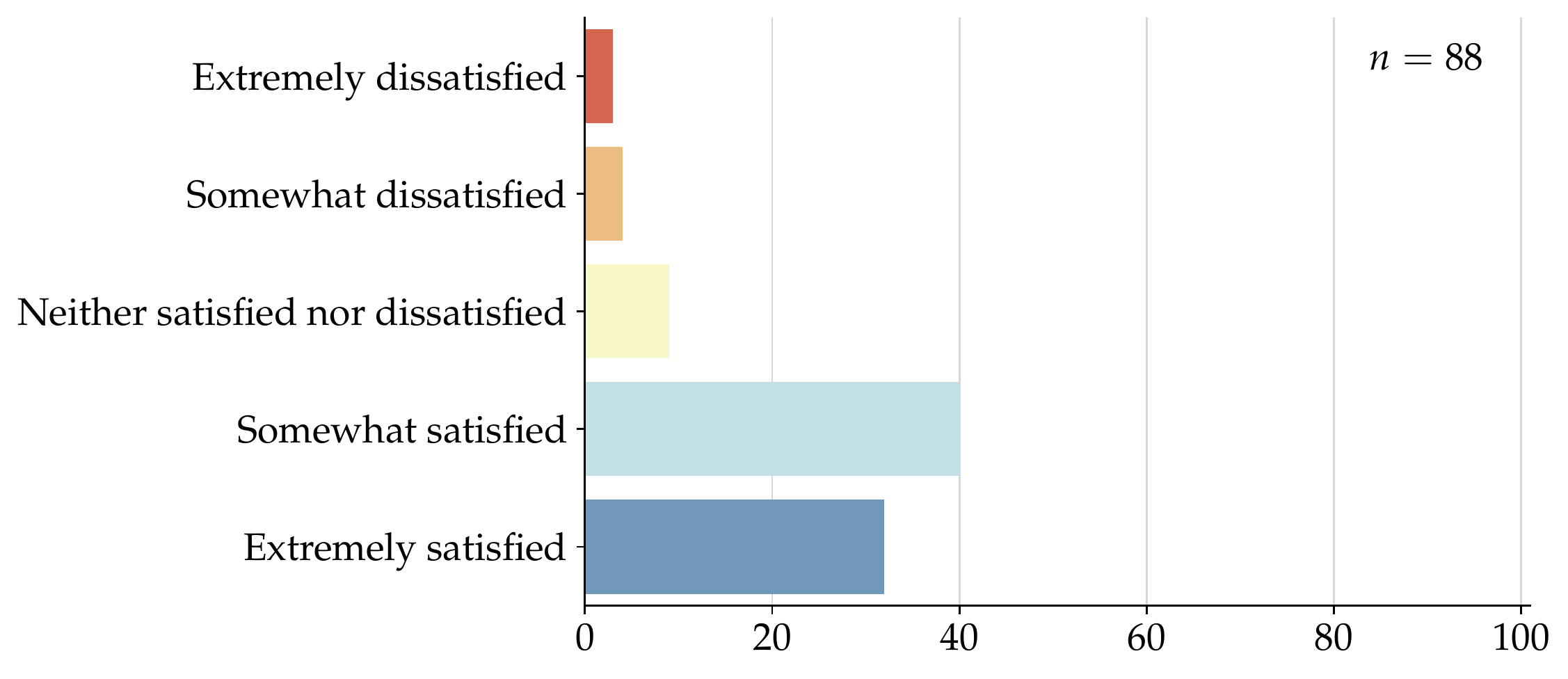}

	\question{As an author, how satisfied are you with the rebuttal phase?}
	
		\marginnote[12pt]{%
		All respondents:\\
		\noindent\includegraphics[width=\linewidth]{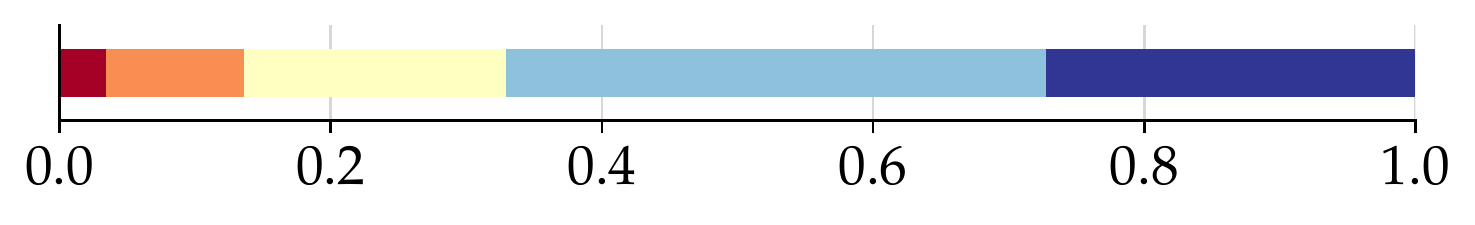}
		
		\noindent Author-only respondents:\\
		\noindent\includegraphics[width=\linewidth]{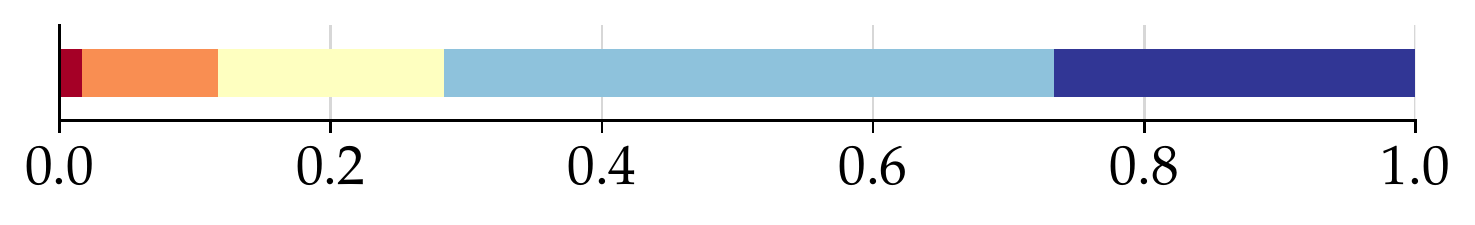}
		
		\noindent Author + \{reviewer, chair\} respondents:\\
		\noindent\includegraphics[width=\linewidth]{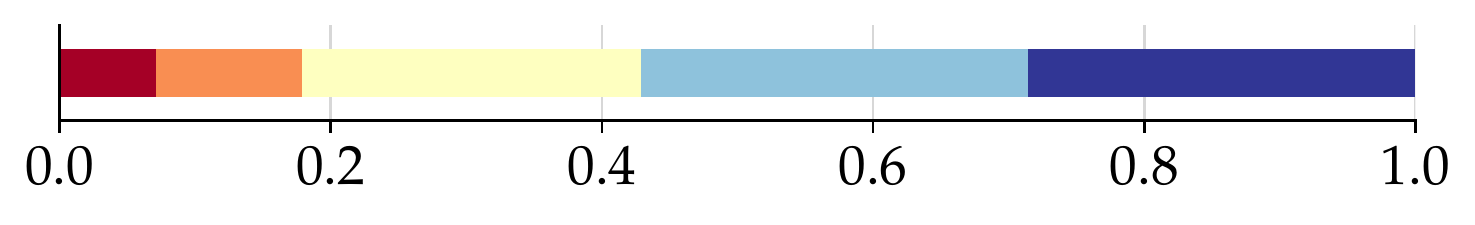}}
	
	\includegraphics[width=\linewidth,keepaspectratio]{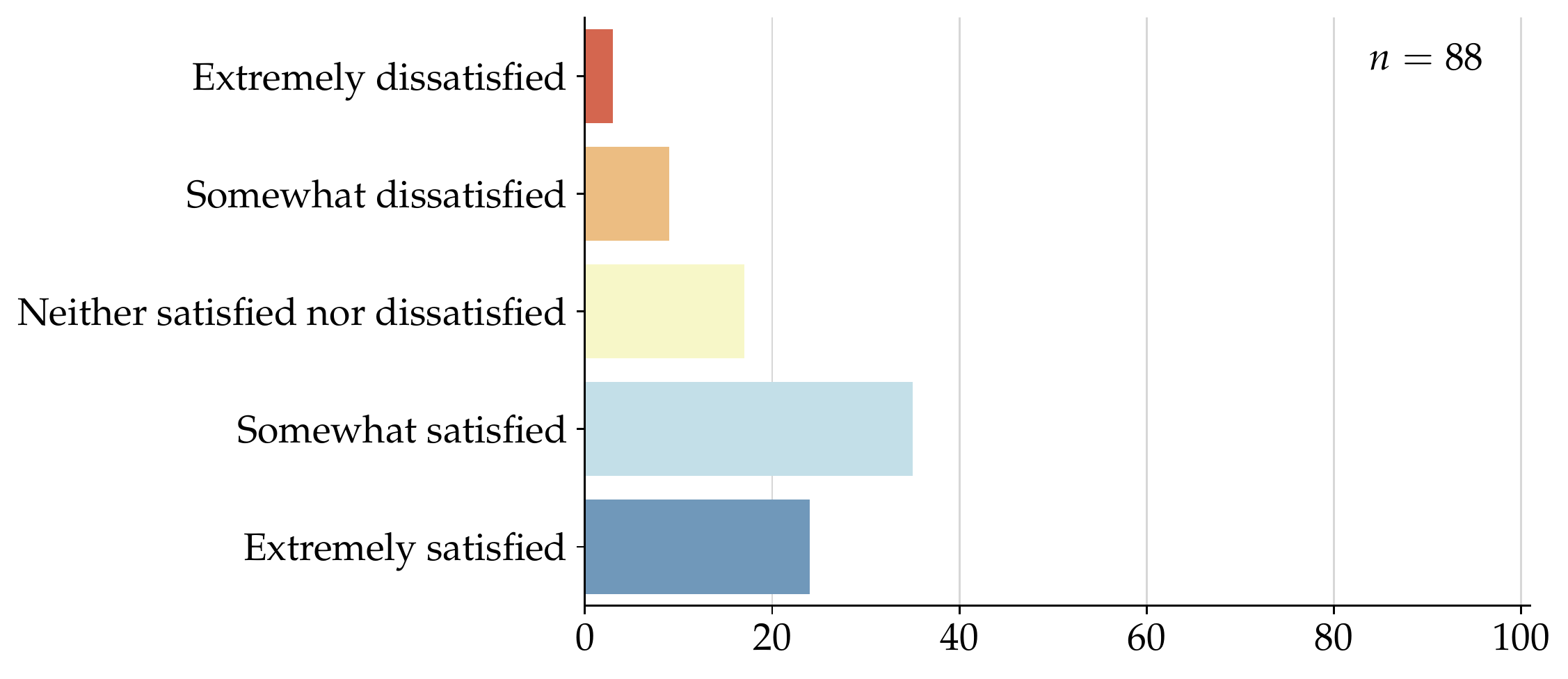}
	
	\question{As an author, how satisfied are you with your review experience overall?}
	
	\marginnote[12pt]{%
		All respondents:\\
		\noindent\includegraphics[width=\linewidth]{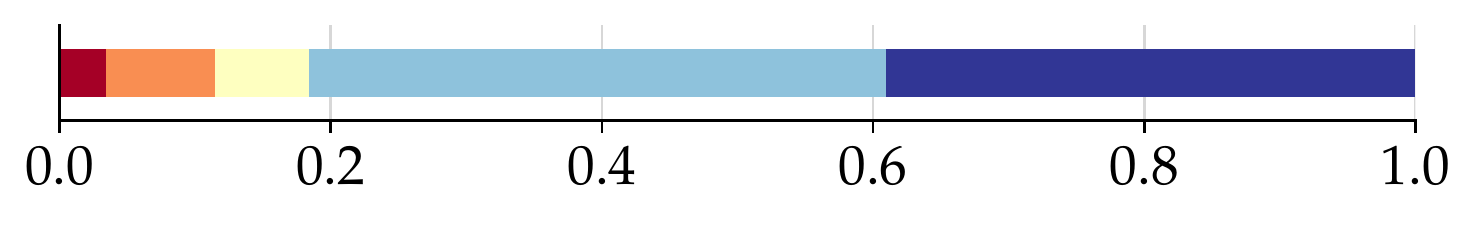}
		
		\noindent Author-only respondents:\\
		\noindent\includegraphics[width=\linewidth]{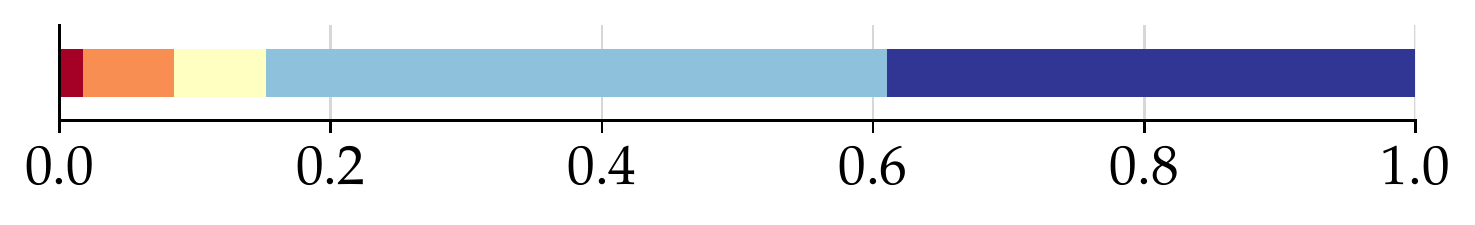}
		
		\noindent Author + \{reviewer, chair\} respondents:\\
		\noindent\includegraphics[width=\linewidth]{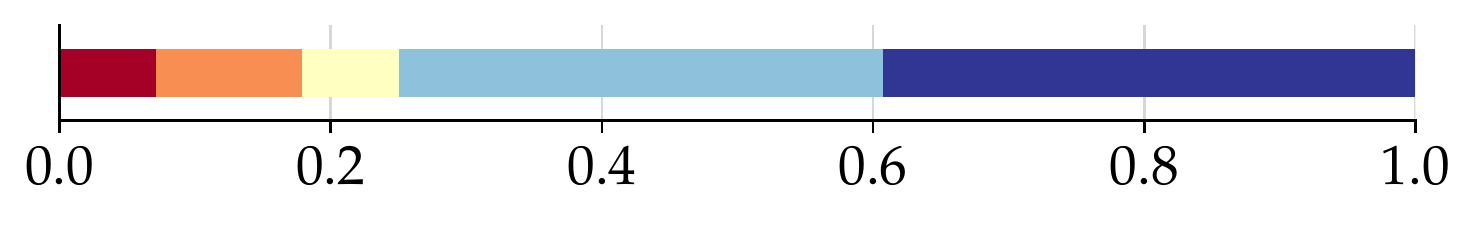}}
	
	\includegraphics[width=\linewidth,keepaspectratio]{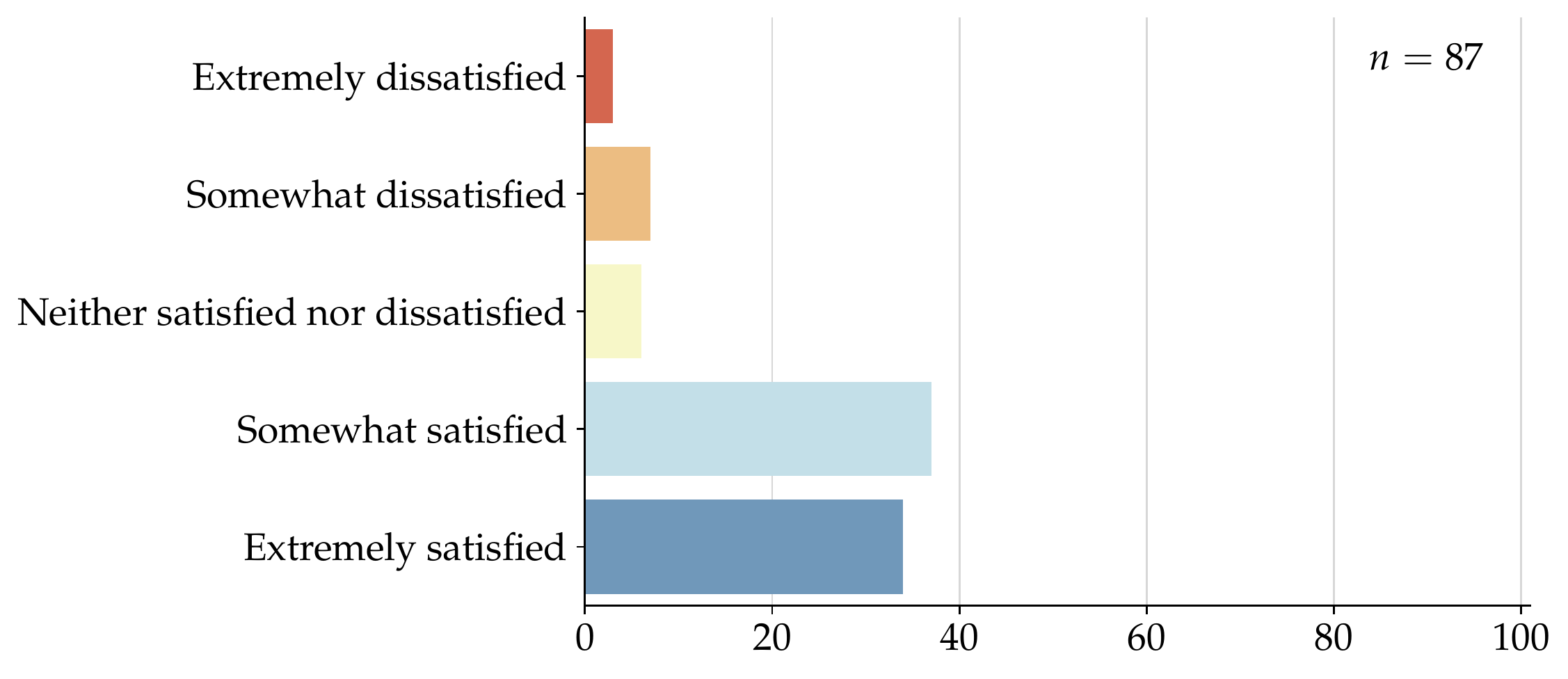}
	
	\clearpage
	
	\question{As an author, how high were the reviewers' standards compared to other AI/ML conferences you submitted to previously?}
	
	\marginnote[12pt]{%
	All respondents:\\
	\noindent\includegraphics[width=\linewidth]{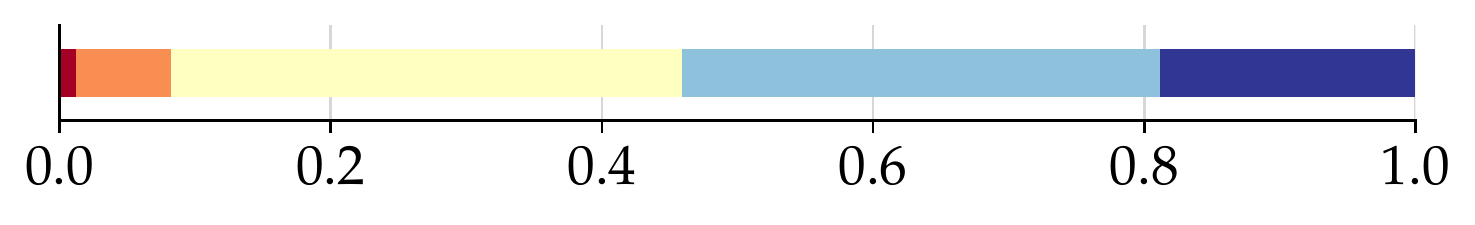}
	
	\noindent Author-only respondents:\\
	\noindent\includegraphics[width=\linewidth]{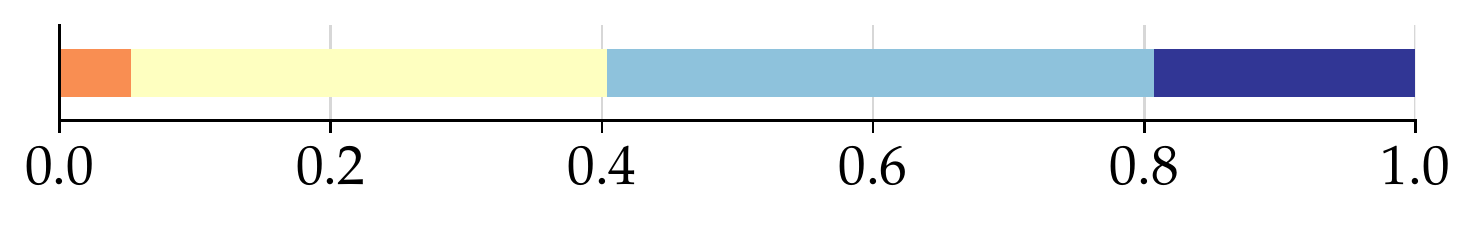}
	
	\noindent Author + \{reviewer, chair\} respondents:\\
	\noindent\includegraphics[width=\linewidth]{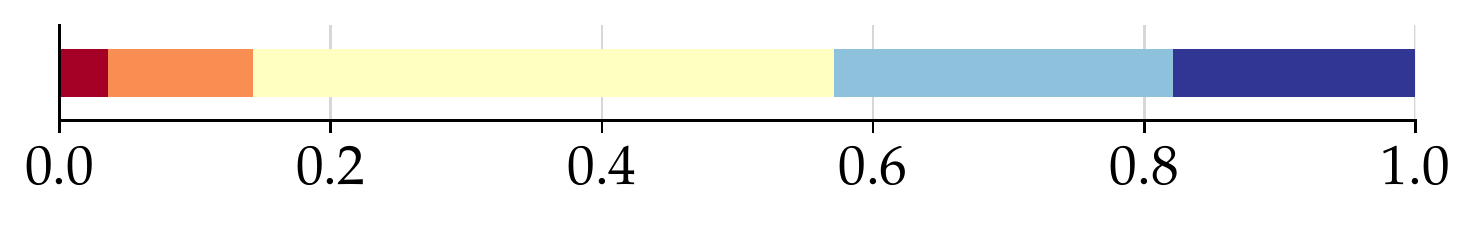}}
	
	\includegraphics[width=\linewidth,keepaspectratio]{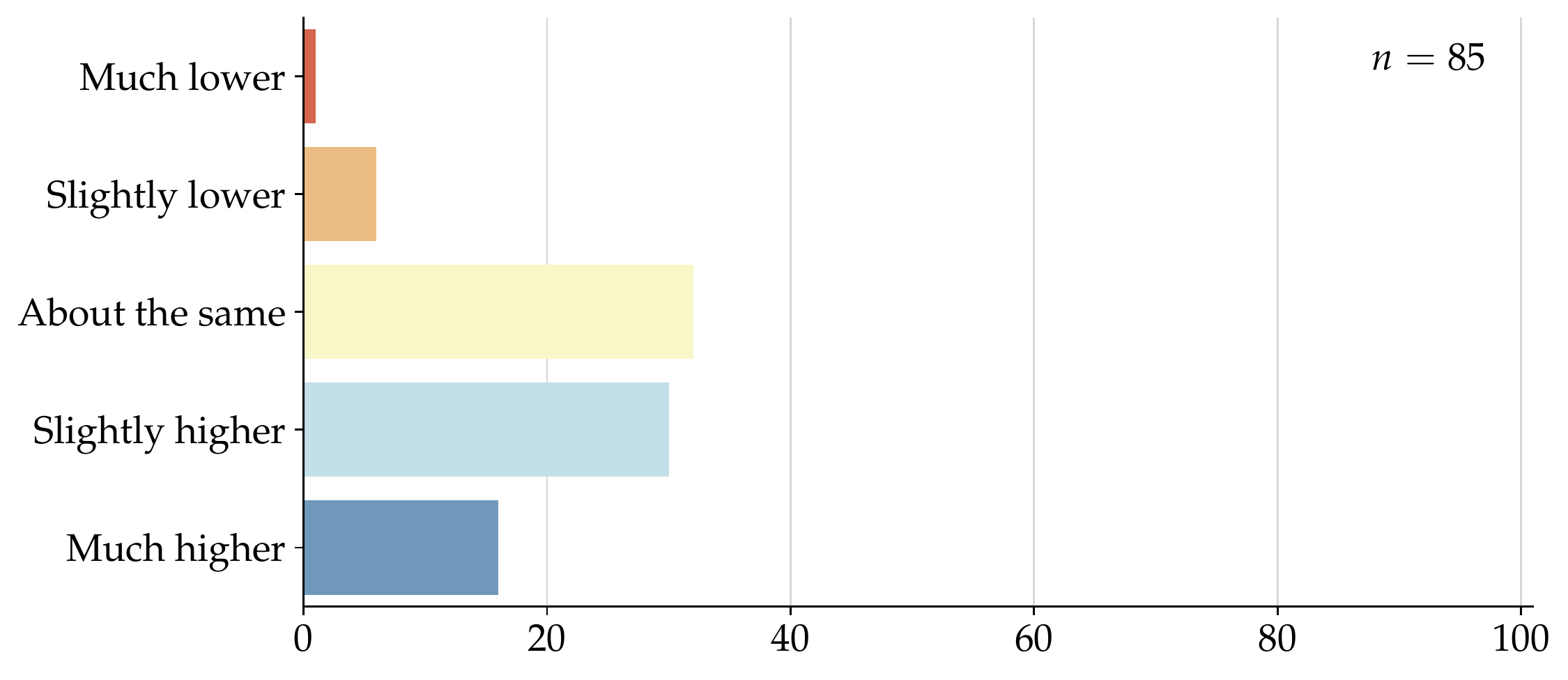}
	
	\question{As an author, how was your review experience compared to other AI/ML conferences you submitted to previously?}
	
	\marginnote[12pt]{%
		All respondents:\\
		\noindent\includegraphics[width=\linewidth]{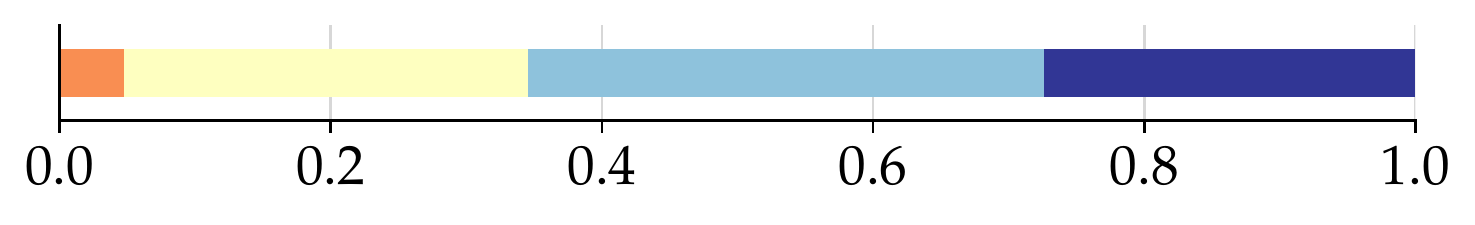}
		
		\noindent Author-only respondents:\\
		\noindent\includegraphics[width=\linewidth]{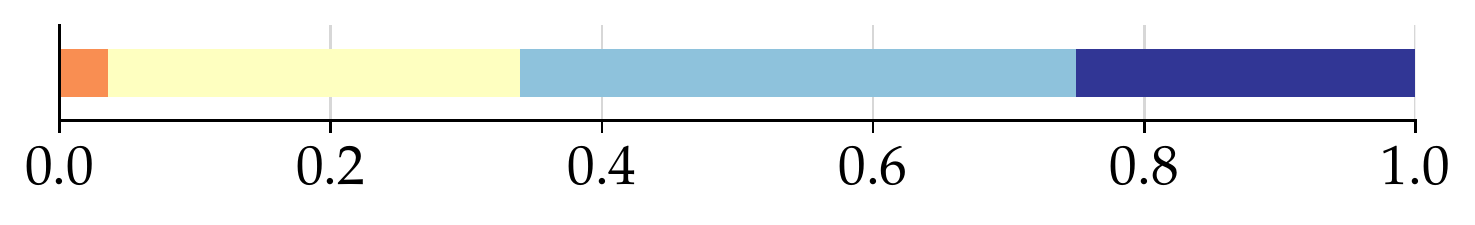}
		
		\noindent Author + \{reviewer, chair\} respondents:\\
		\noindent\includegraphics[width=\linewidth]{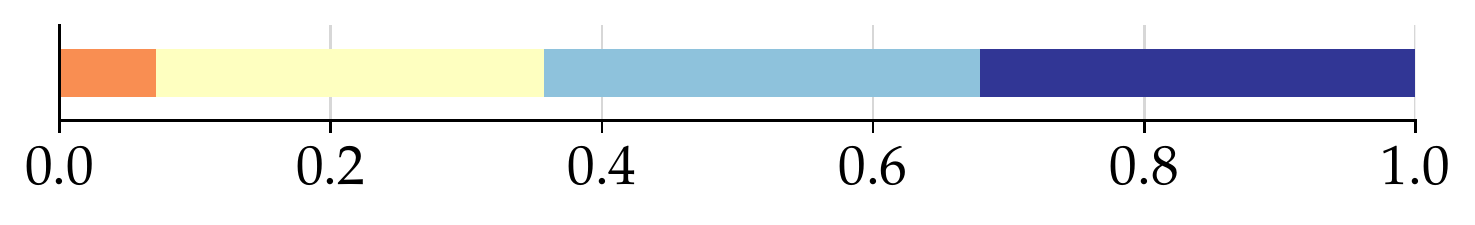}}
	
	\includegraphics[width=\linewidth,keepaspectratio]{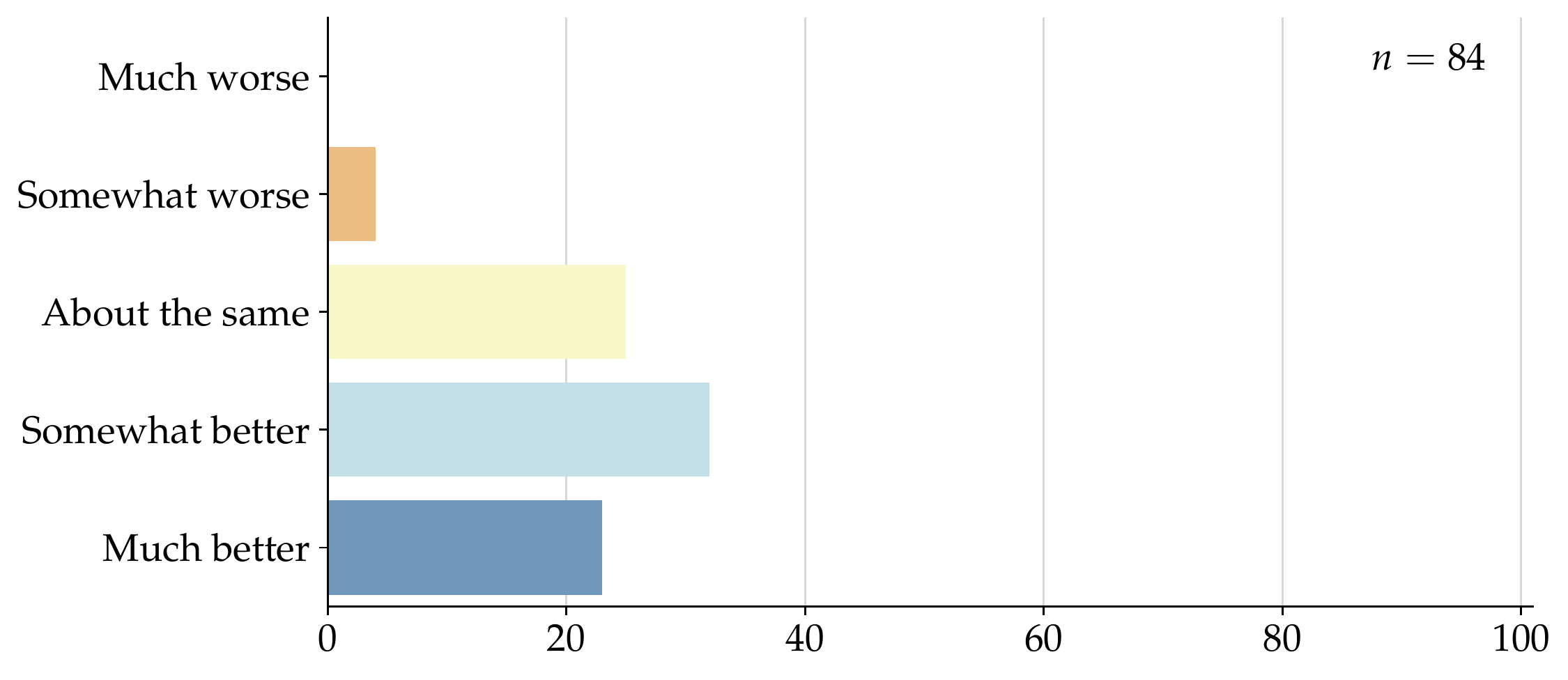}

	\subsection{Questions to Reviewers}

  {
    \emergencystretch 3em
    We find that interest in the conference topic is the factor most
    frequently mentioned as a motivation to review for \LoG,
    followed by the prospect of a monetary reward, and being
    asked to review based on one's own professional network. Moreover,
    reviewers are moderately satisfied with the rebuttal phase and the
    review experience overall, mentioning that their experience is
    comparable to those of more established conferences. More than 50\% of
    the reviewers also report that their review load was slightly lower or
    much lower in comparison to other conferences. Given that the program
    committee assigned virtually all reviewers no more than~$3$ papers~(with few
    exceptions for certain expert and emergency reviewers, who were
    assigned up to~$5$ papers), the feedback provides a good justification
    for continuing to keep the review load low. 
  }
  
\clearpage

	\question{As a reviewer, why did you choose to review for the conference?}
	
	\includegraphics[width=\linewidth,keepaspectratio]{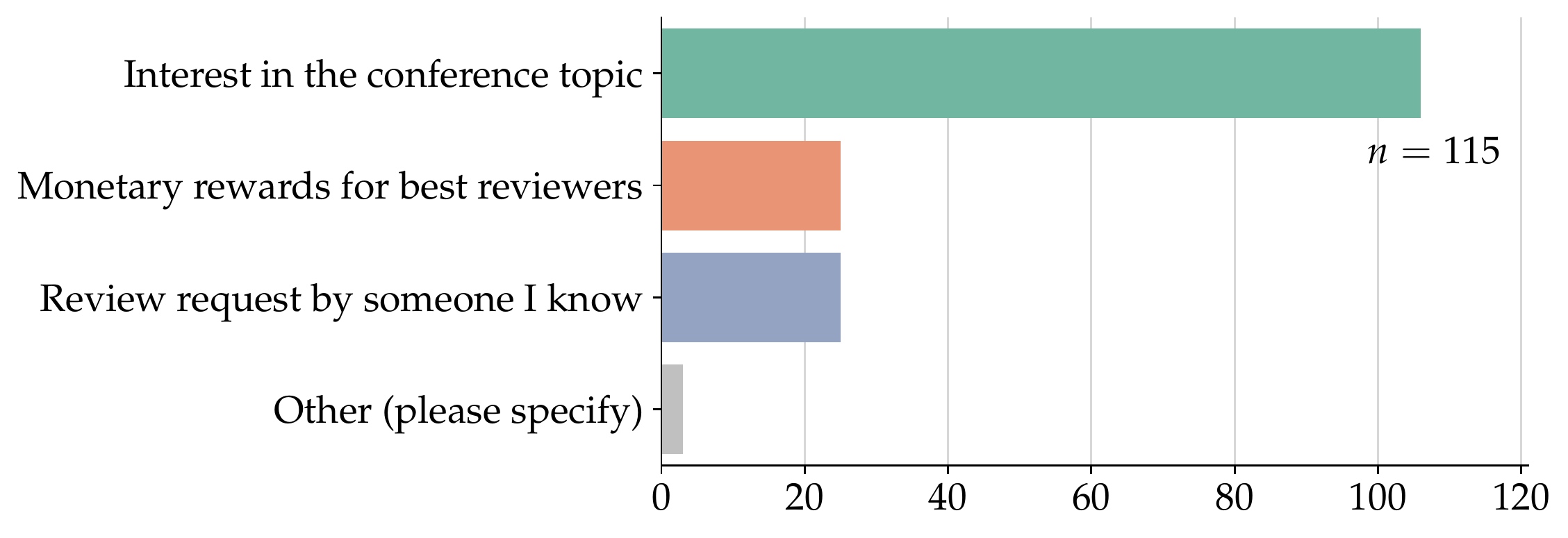}
	
	\noindent All reviewers who indicated that they were motivated by the monetary rewards offered to best reviewers 
	also indicated that they were motivated by at least one other factor:

	\includegraphics[width=\linewidth,keepaspectratio]{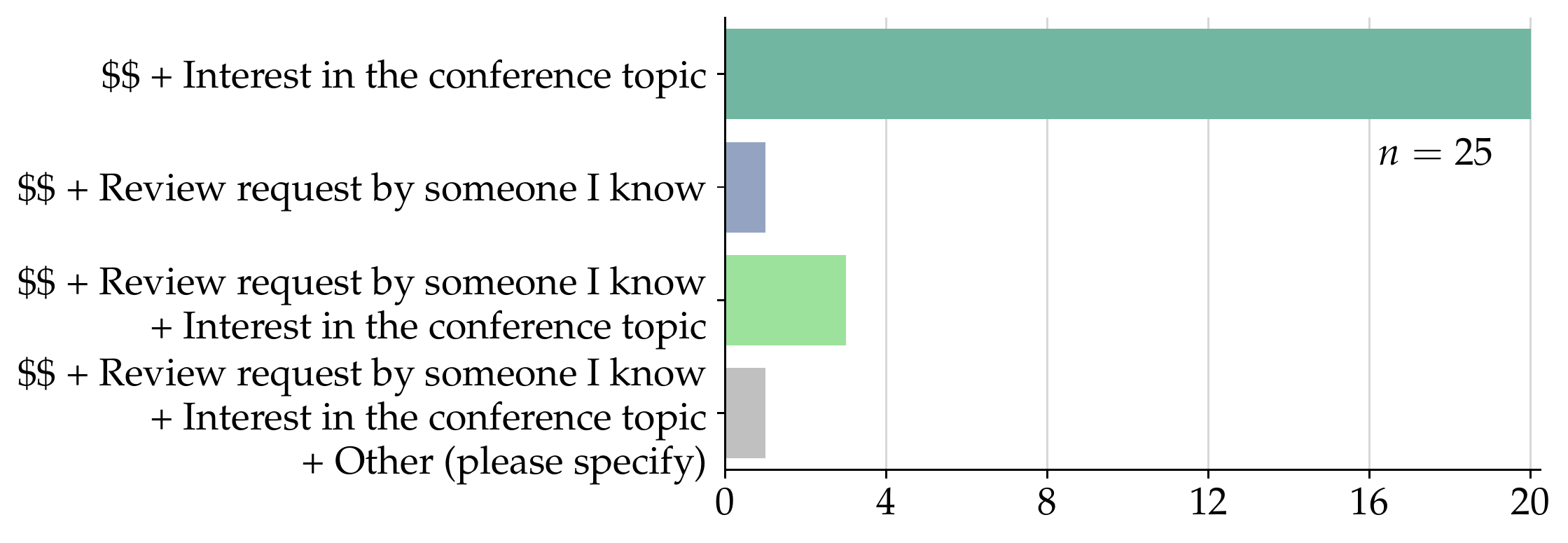}

  \noindent Among the responses given in text form, reviewers stated that they
  were either recruited as an emergency reviewer~($n = 1$) or chose \LoG
  because of the reputation of the organizers~($n = 1$).
	
	\question{As a reviewer, how satisfied are you with the rebuttal phase?}
	\marginnote[12pt]{%
		All respondents:\\
		\noindent\includegraphics[width=\linewidth]{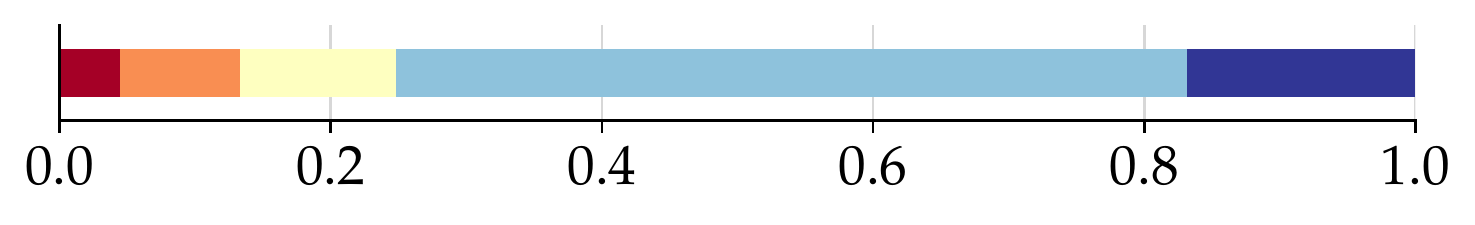}
		
		\noindent Reviewer-only respondents:\\
		\noindent\includegraphics[width=\linewidth]{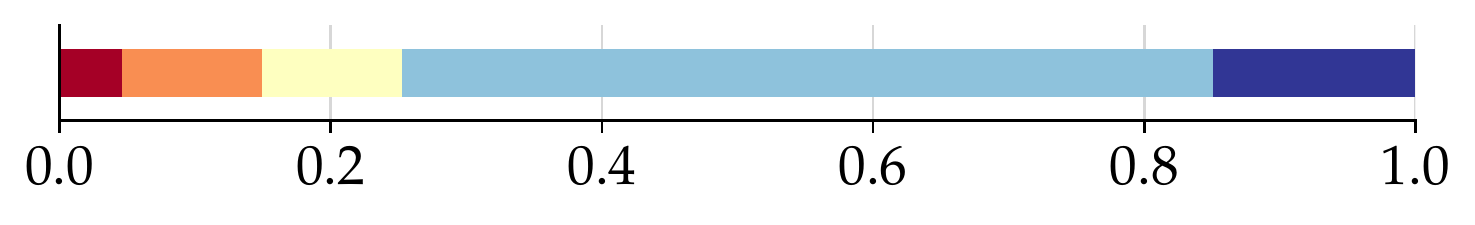}
		
		\noindent Reviewer + author respondents:\\
		\noindent\includegraphics[width=\linewidth]{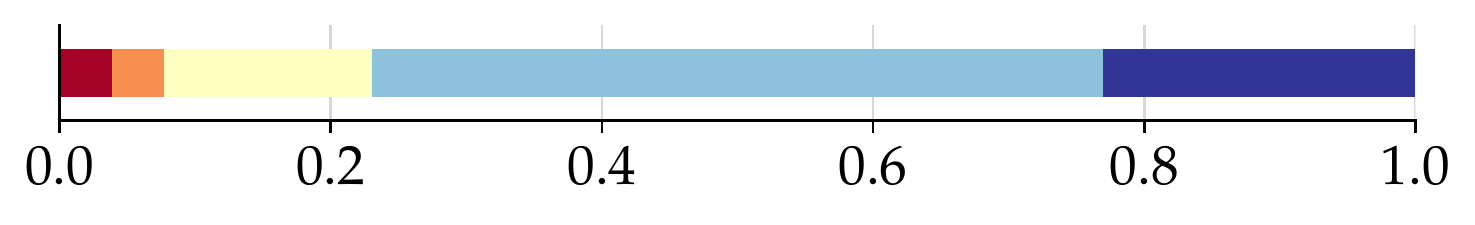}}
	
	\includegraphics[width=\linewidth,keepaspectratio]{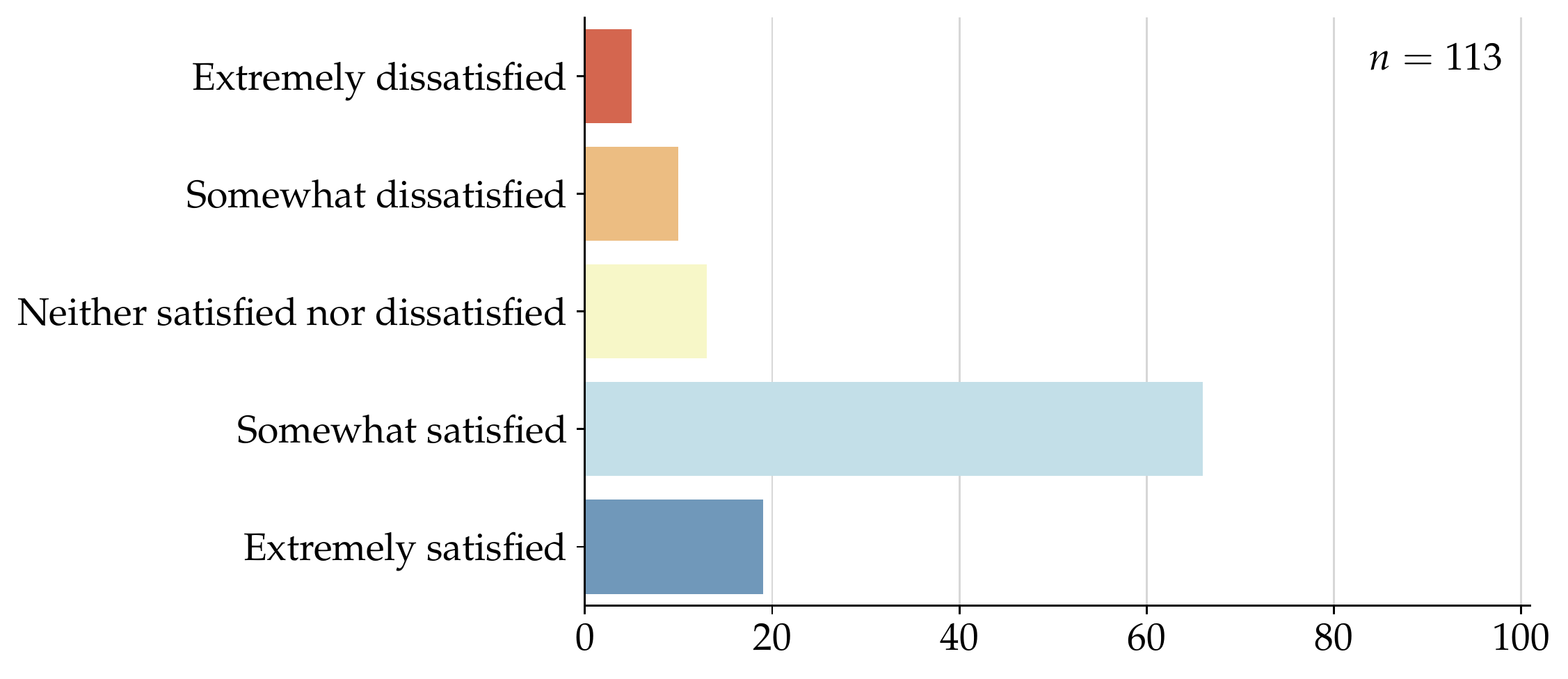}
	
	\clearpage
	
	\question{As a reviewer, how satisfied are you with your review experience overall?}
	\marginnote[12pt]{%
		All respondents:\\
		\noindent\includegraphics[width=\linewidth]{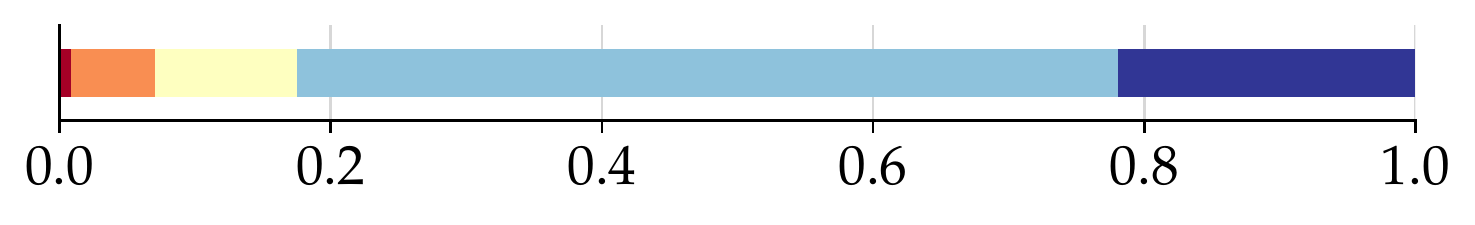}
		
		\noindent Reviewer-only respondents:\\
		\noindent\includegraphics[width=\linewidth]{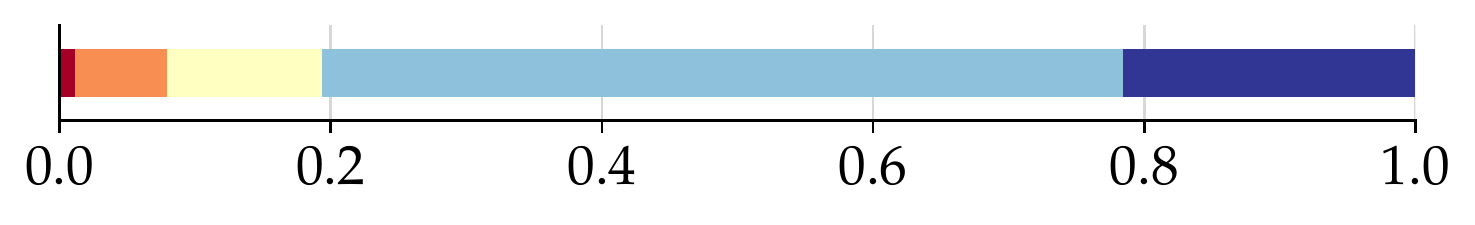}
		
		\noindent Reviewer + author respondents:\\
		\noindent\includegraphics[width=\linewidth]{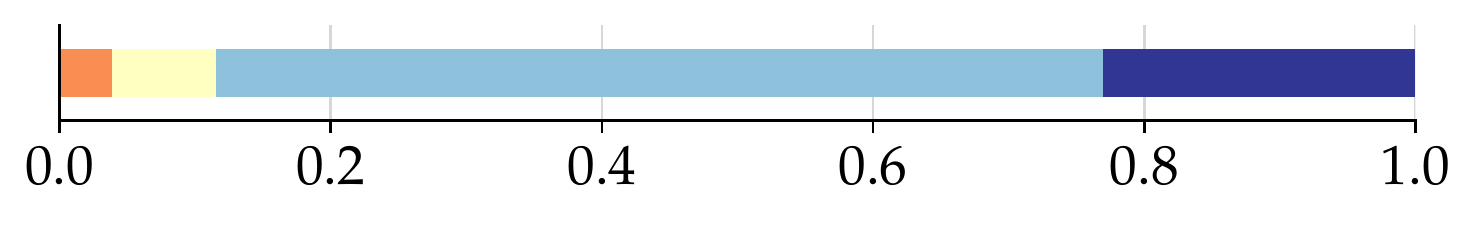}}
	
	\includegraphics[width=\linewidth,keepaspectratio]{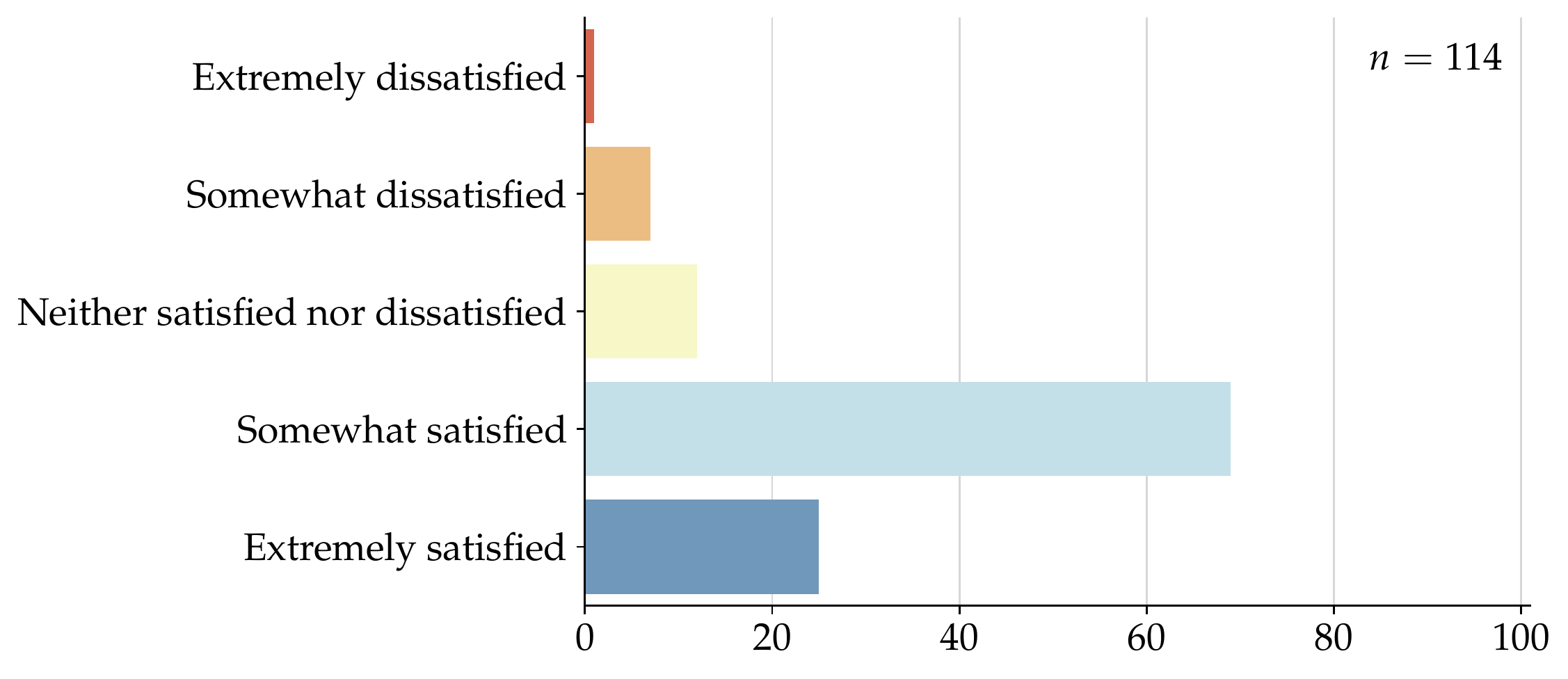}
	
	\question{As a reviewer, how was your review experience compared to other AI/ML conferences you reviewed for previously?}
	\marginnote[12pt]{%
		All respondents:\\
		\noindent\includegraphics[width=\linewidth]{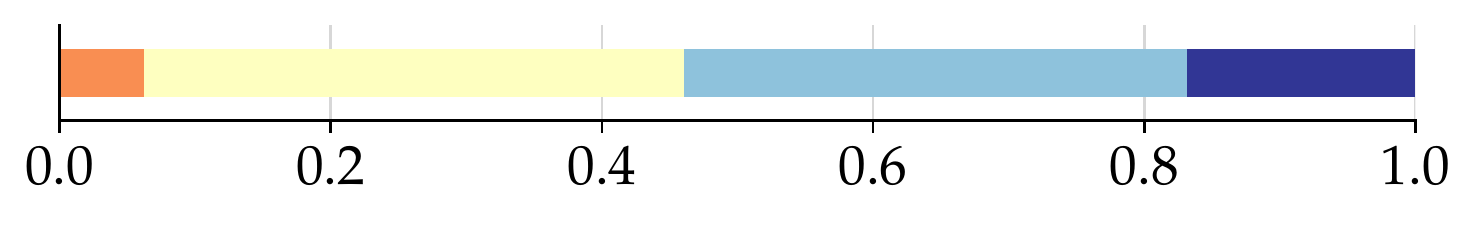}
		
		\noindent Reviewer-only respondents:\\
		\noindent\includegraphics[width=\linewidth]{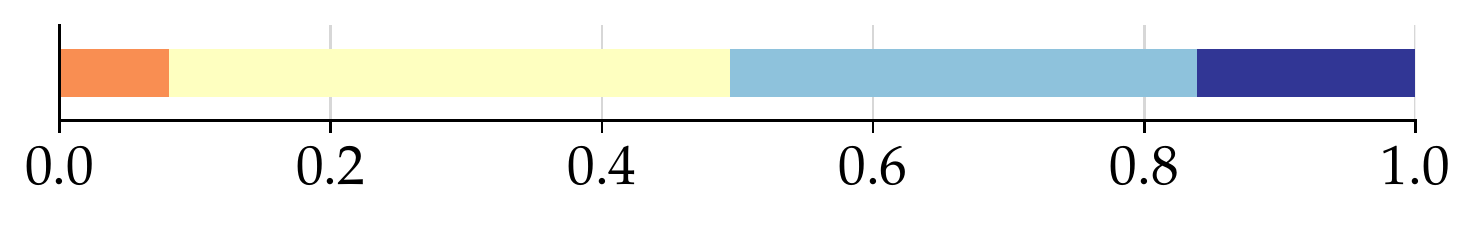}
		
		\noindent Reviewer + author respondents:\\
		\noindent\includegraphics[width=\linewidth]{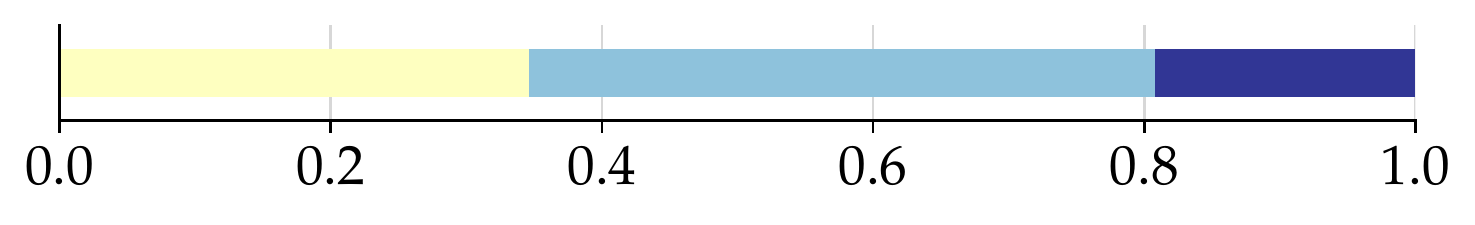}}
	
	\includegraphics[width=\linewidth,keepaspectratio]{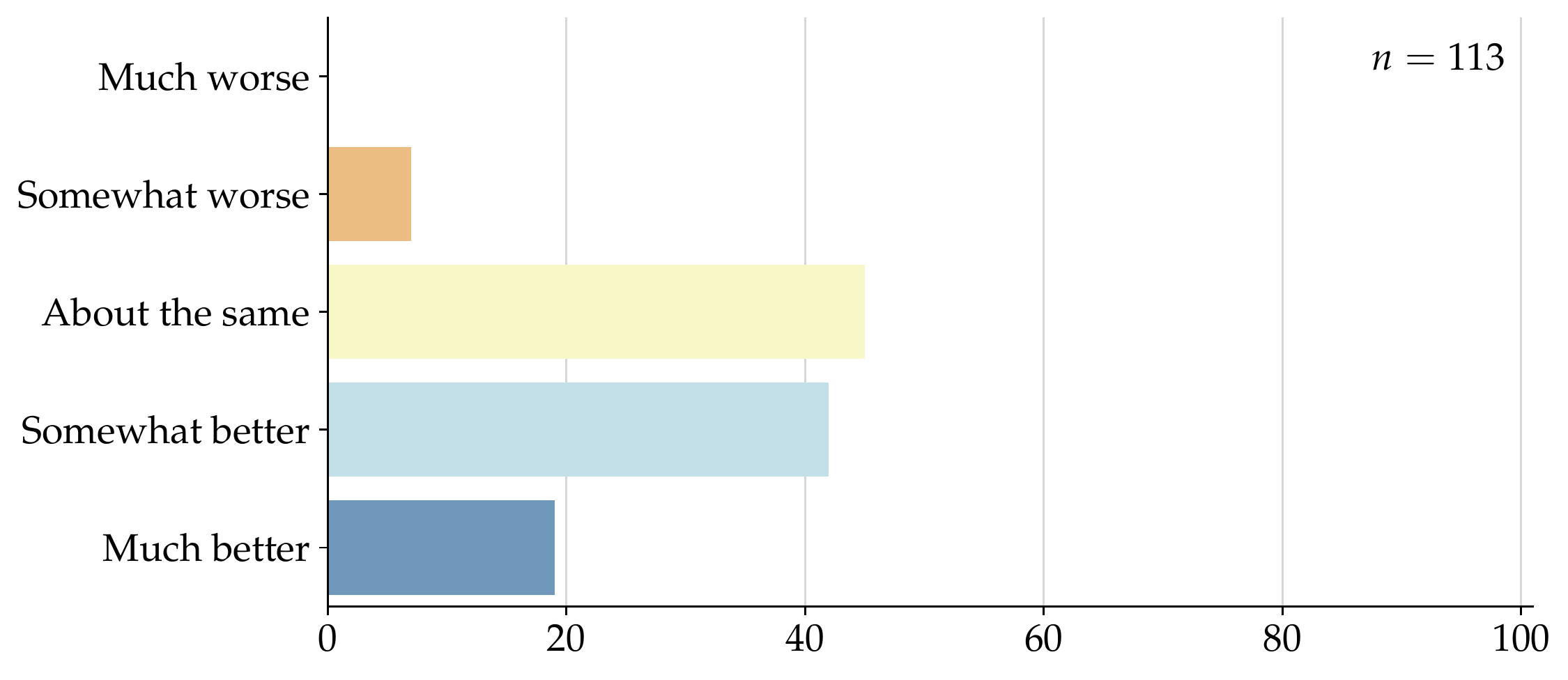}
	
	\question{As a reviewer, how was your workload compared to other AI/ML conferences you reviewed for previously?}
	\marginnote[12pt]{%
		All respondents:\\
		\noindent\includegraphics[width=\linewidth]{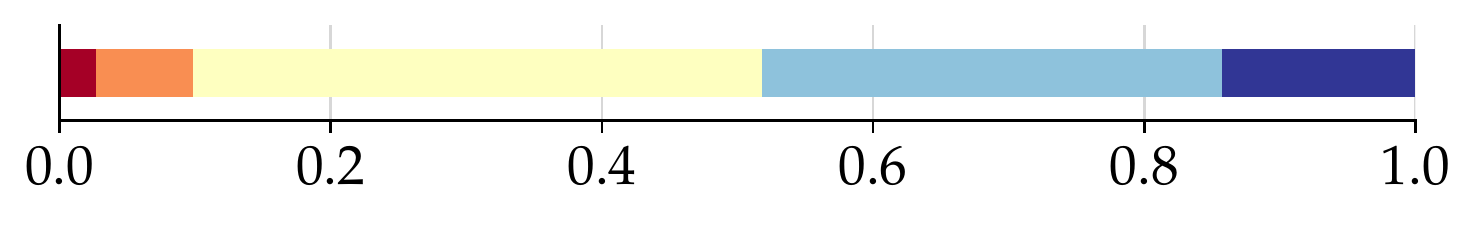}
		
		\noindent Reviewer-only respondents:\\
		\noindent\includegraphics[width=\linewidth]{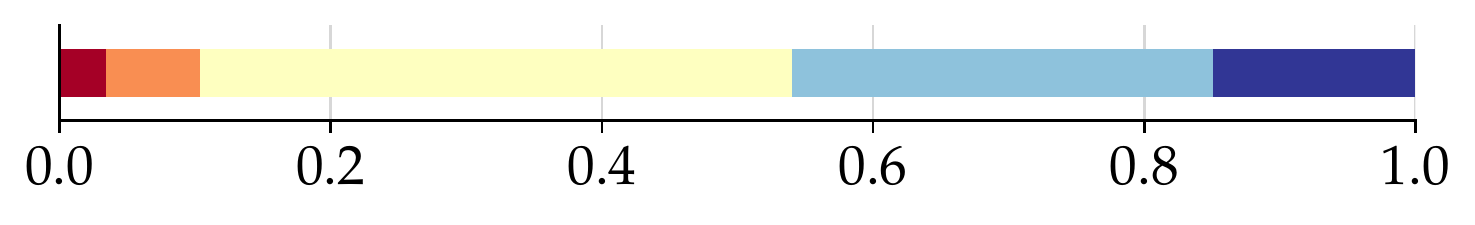}
		
		\noindent Reviewer + author respondents:\\
		\noindent\includegraphics[width=\linewidth]{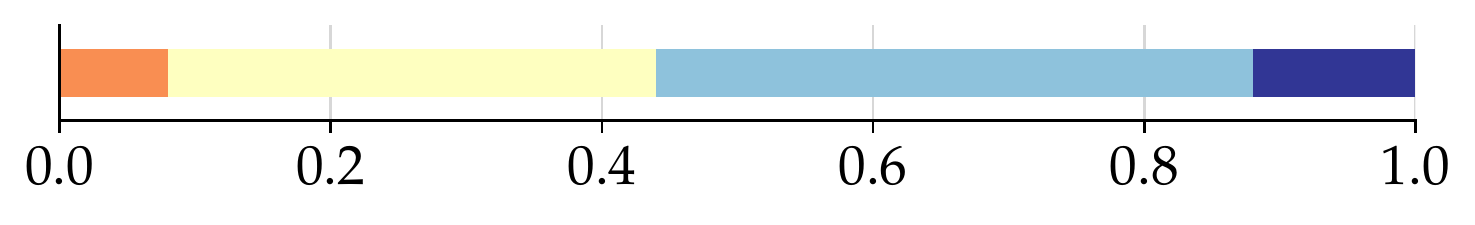}}
	
	\includegraphics[width=\linewidth,keepaspectratio]{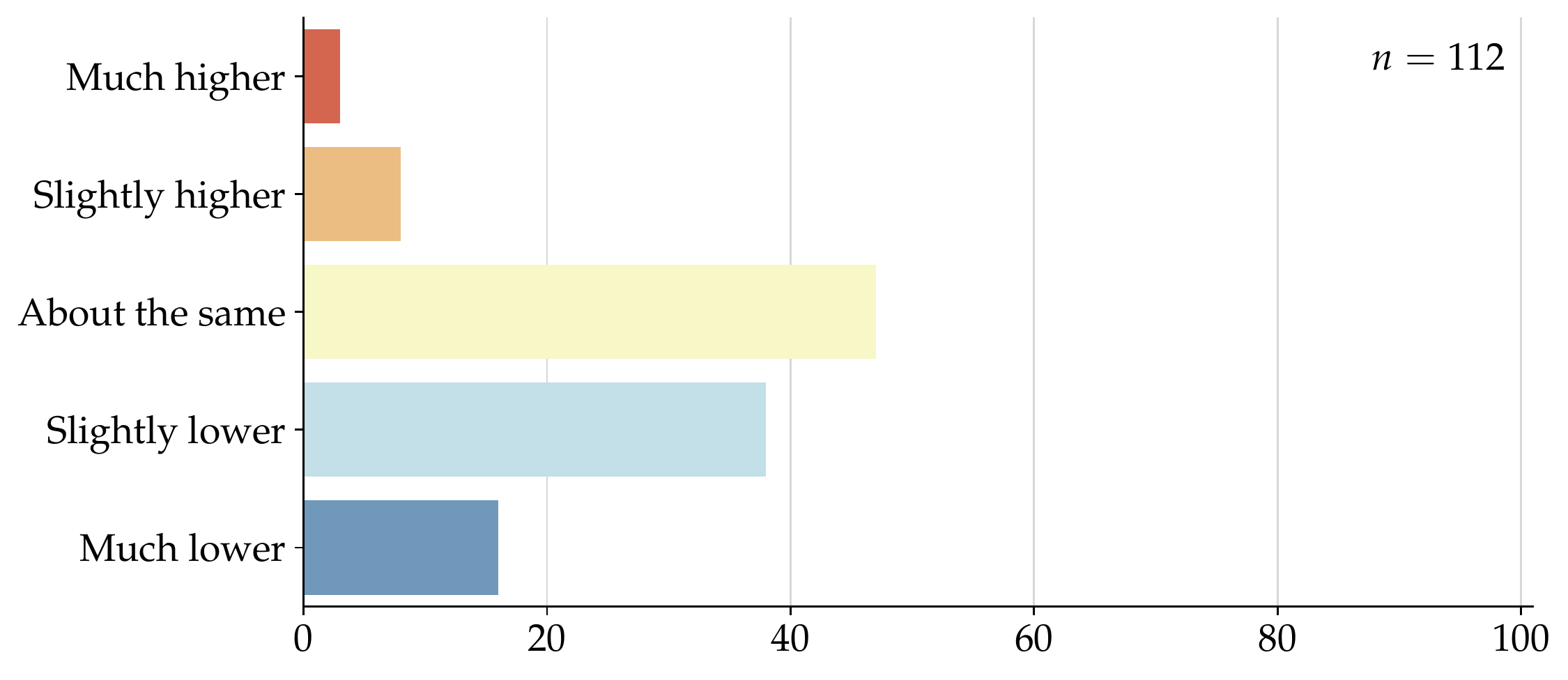}
	
	\clearpage
	
	\noindent About 30\% of responding reviewers indicated that they reviewed for ICLR 2023 while reviewing for \LoG, with the median affirmative respondent reviewing 5--6 papers for ICLR 2023. 
	Most affected reviewers were neutral or critical toward their double duty. 
	This highlights the importance of conference timing when calibrating reviewer workloads.
	
	\question{Did you review for ICLR 2023?}
	
	\marginnote[12pt]{%
		All respondents:\\
		\noindent\includegraphics[width=\linewidth]{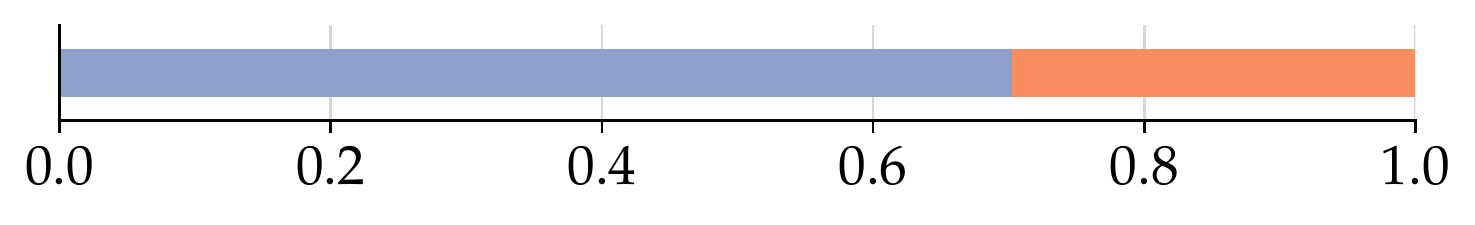}
		
		\noindent Reviewer-only respondents:\\
		\noindent\includegraphics[width=\linewidth]{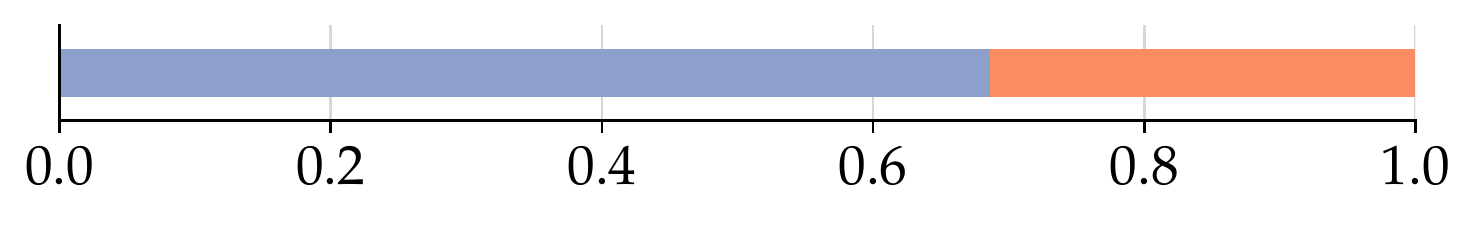}
		
		\noindent Reviewer + author respondents:\\
		\noindent\includegraphics[width=\linewidth]{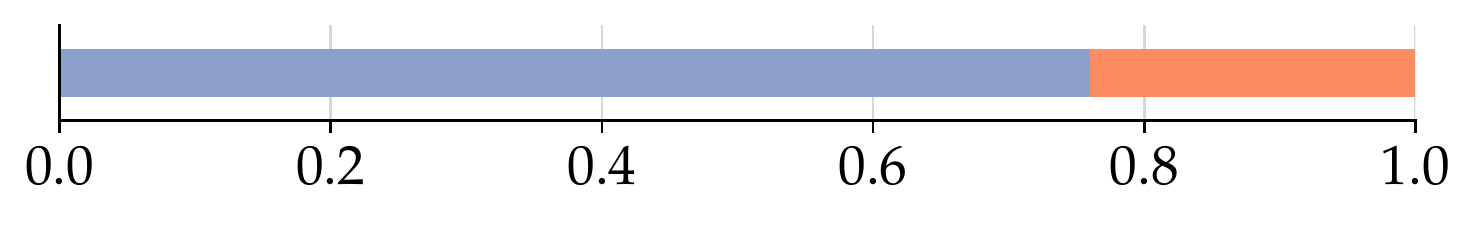}}
	
	\includegraphics[width=\linewidth,keepaspectratio]{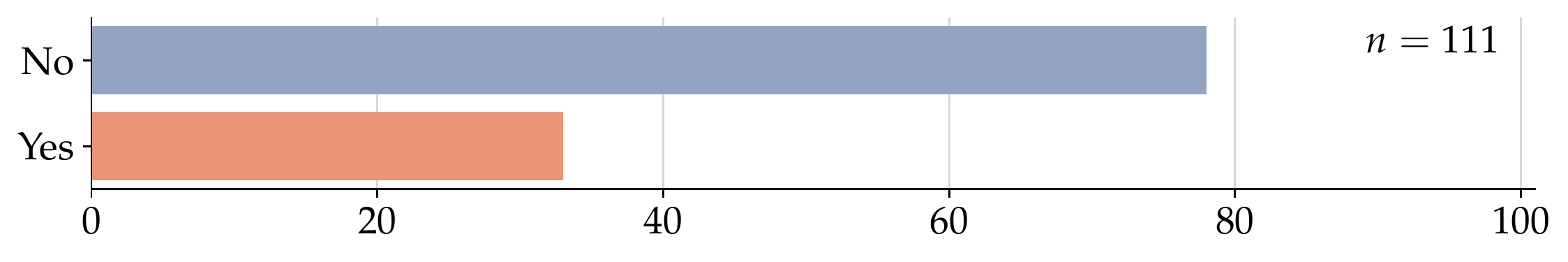}
	
	\vspace*{48pt}
	
	\question{How many papers did you review for ICLR 2023?}
	\marginnote[12pt]{%
		All respondents:\\
		\noindent\includegraphics[width=\linewidth]{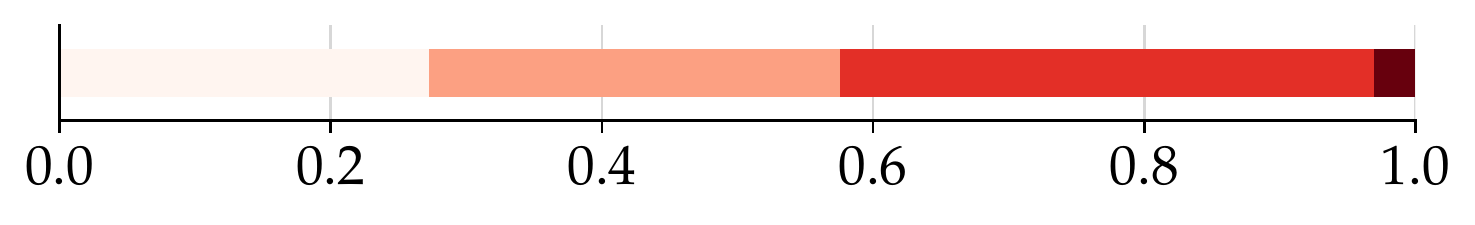}
		
		\noindent Reviewer-only respondents:\\
		\noindent\includegraphics[width=\linewidth]{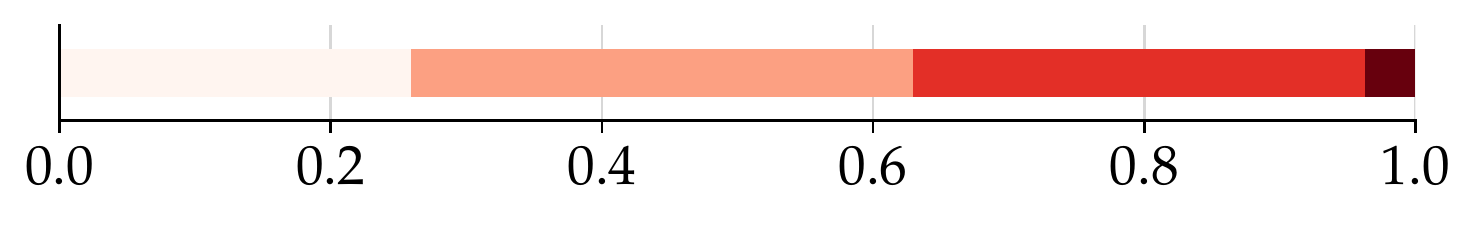}
		
		\noindent Reviewer + author respondents:\\
		\noindent\includegraphics[width=\linewidth]{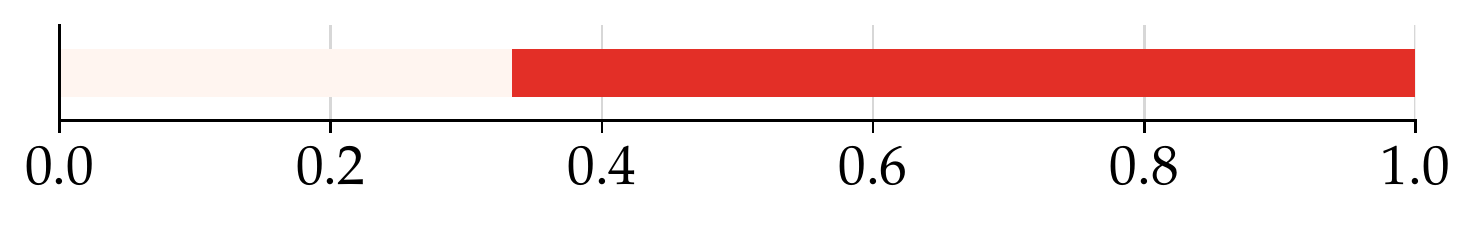}}
	
	\includegraphics[width=\linewidth,keepaspectratio]{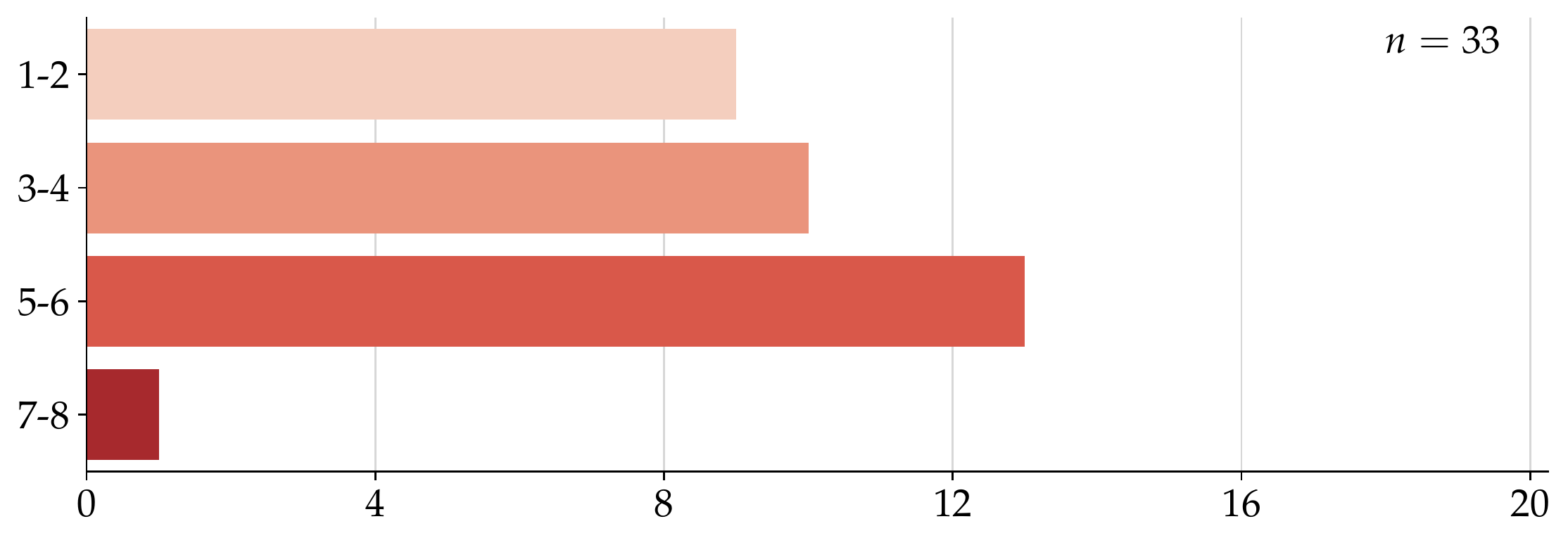}

	\question{If you reviewed for ICLR 2023: How do you feel about reviewing for ICLR and LoG at the same time?}
	\marginnote[12pt]{%
		All respondents:\\
		\noindent\includegraphics[width=\linewidth]{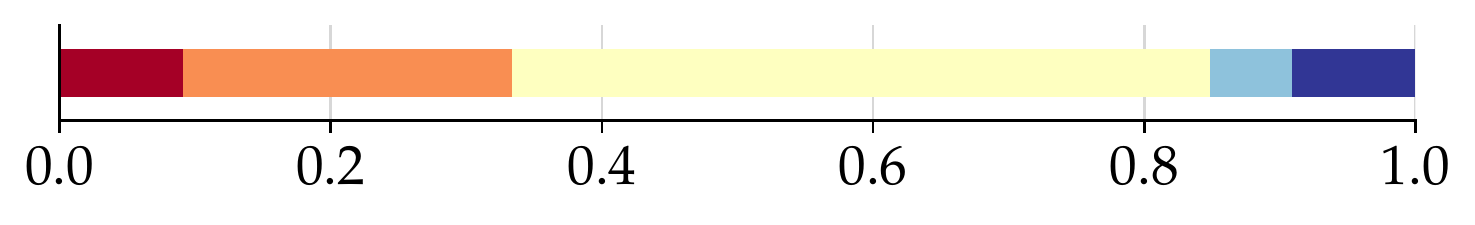}
		
		\noindent Reviewer-only respondents:\\
		\noindent\includegraphics[width=\linewidth]{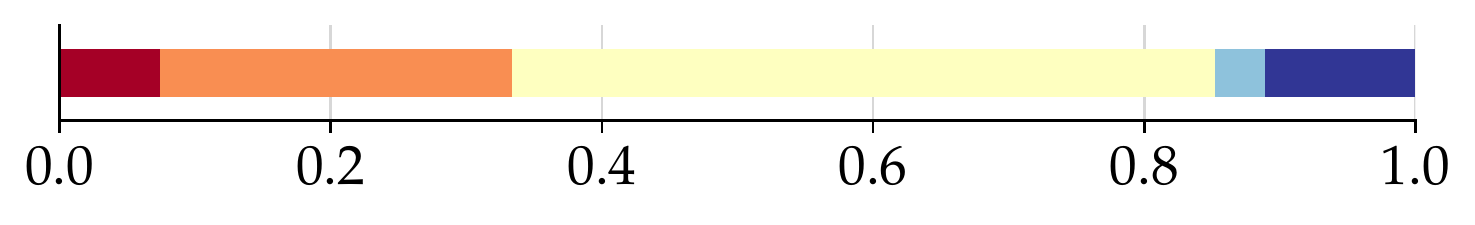}
		
		\noindent Reviewer + author respondents:\\
		\noindent\includegraphics[width=\linewidth]{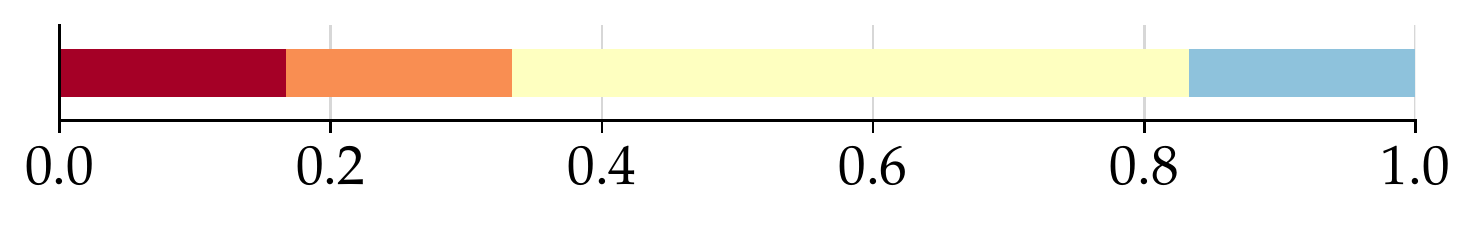}}
	
	\includegraphics[width=\linewidth,keepaspectratio]{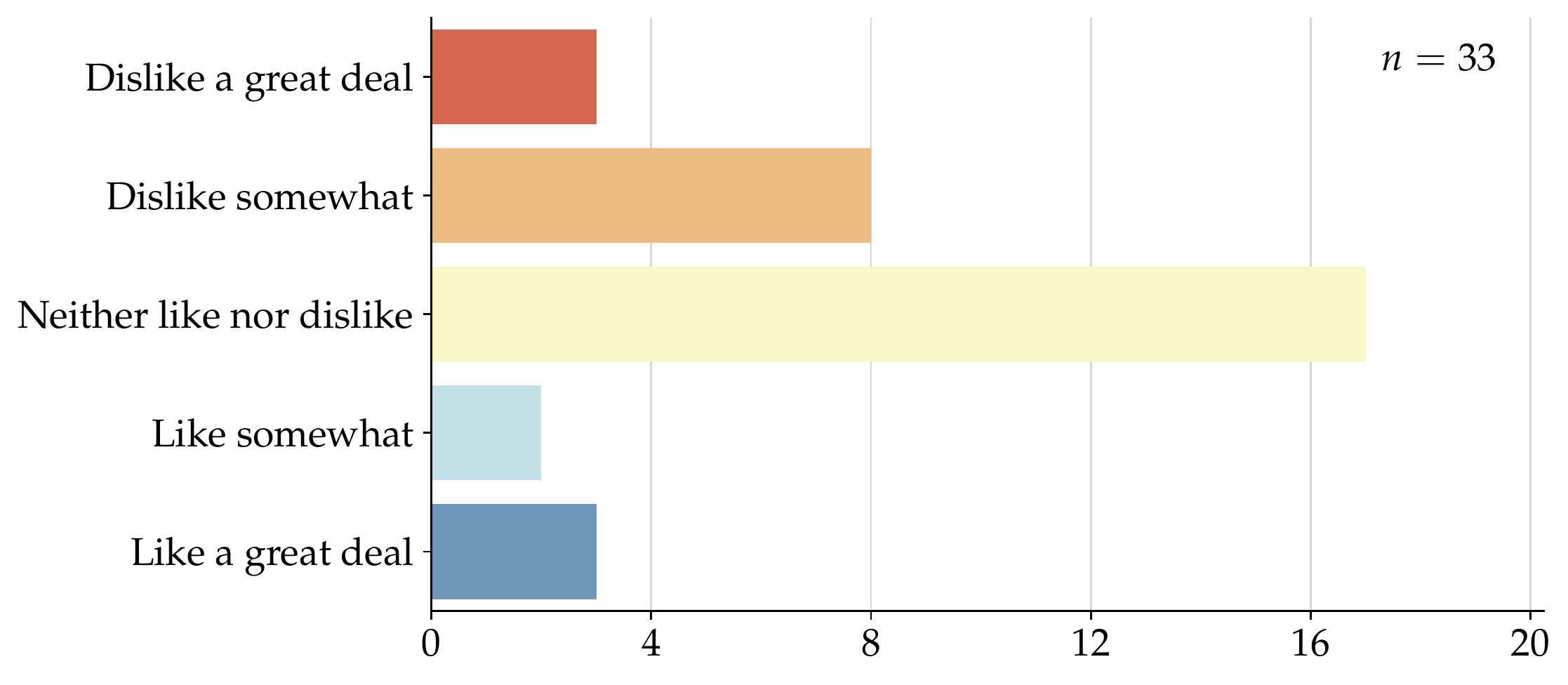}
	
	\clearpage
	
	\subsection{Questions to Area Chairs}
	
	We refrain from illustrating area chairs' responses due to their small number ($n=3$).
	
	\question{As an area chair, how satisfied are you with your review experience overall?}
	Responding area chairs were somewhat or extremely satisfied with their review experience overall.
	
	\question{As an area chair, how was your review experience compared to other AI/ML conferences you reviewed for previously?}
	Responding area chairs judged their review experience to be much better or about the same as for other AI/ML conferences they reviewed for previously.
	
	\question{As an area chair, how was your workload compared to other AI/ML conferences you reviewed for previously?}
	Responding area chairs judged their workload to be slightly lower than or about the same as for other AI/ML conferences they reviewed for previously.
	
	\subsection{General Questions}

  We also gave participants the option to answer questions about the
  current setup of the conference~(one track for full papers and one track for extended
  abstracts) and provided options for free-form feedback. The latter received $n = 51$ responses, which we summarize below.
	
  \question{How could we improve your review experience?}

  In general, most respondents wanted to have more time to discuss their papers
  with reviewers and mentioned that reviewers should be encouraged to
  be more \emph{active} during the rebuttal phase. Some commenters
  raised unreasonable demands by reviewers, such as irrelevant
  experiments and out-of-scope citations, as a prevailing issue of
  machine learning conferences that they did \emph{not} experience with
  \LoG. A prevalent wish was also to enable rating of reviewers by authors, as
  well as to establish a better culture of reviewing that moves away
  from mere numerical scores. Paraphrasing the respondents here, there
  appears to be a call for more nuance in the reviewing process.
  Interestingly, several respondents strongly suggested the utility of
  enabling public comments on submissions to engage the community in the
  reviewing process.
  Finally, some commenters took the time to remark that their experience
  stood out in positive terms when compared to other conferences.

  Concerning the different tracks, respondents commented that the
  separation should be explained better to authors and reviewers
  alike.
  With reviewers having similarly high standards for work that is
  clearly still in progress, getting an extended abstract accepted was
  perceived as a tough challenge for authors.

	\question{How do you like the ``Extended Abstract'' track?}
	\marginnote[12pt]{%
		All respondents:\\
		\noindent\includegraphics[width=\linewidth]{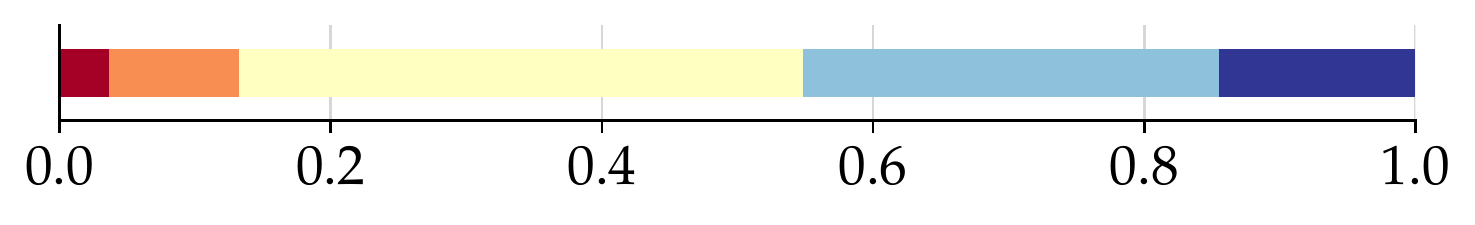}
		
		\noindent Author-only respondents:\\
		\noindent\includegraphics[width=\linewidth]{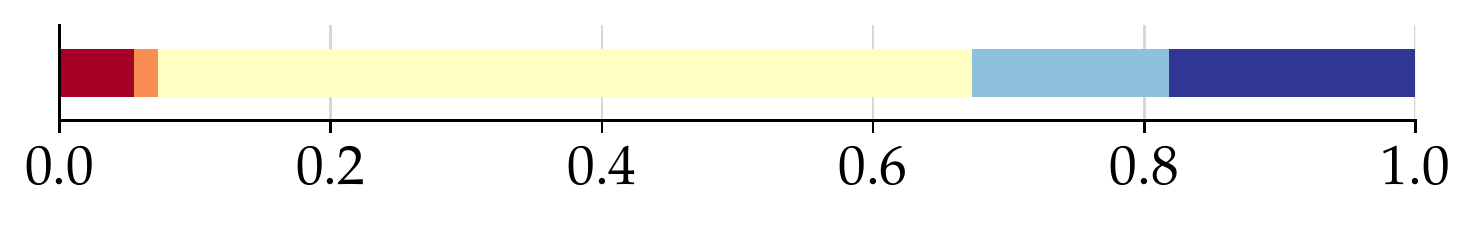}
		
		\noindent Reviewer-only respondents:\\
		\noindent\includegraphics[width=\linewidth]{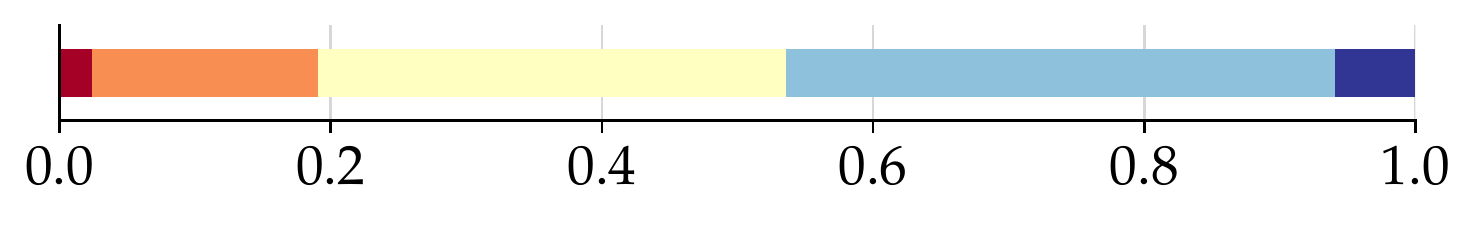}
		
		\noindent Author + \{reviewer, chair\} respondents:\\
		\noindent\includegraphics[width=\linewidth]{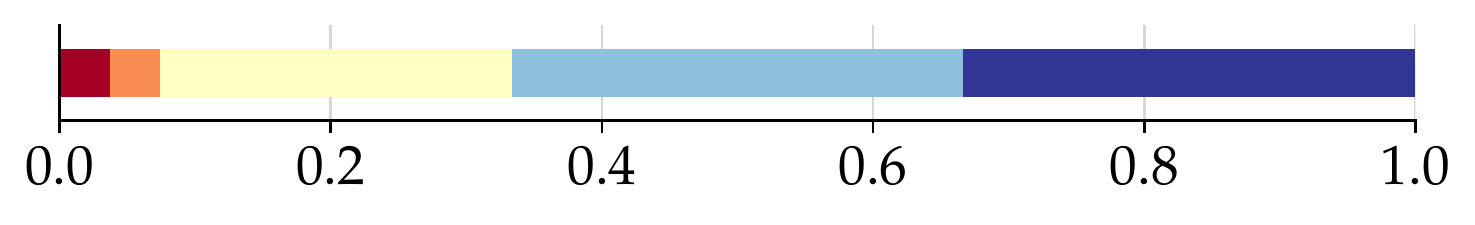}}
	
	\includegraphics[width=\linewidth,keepaspectratio]{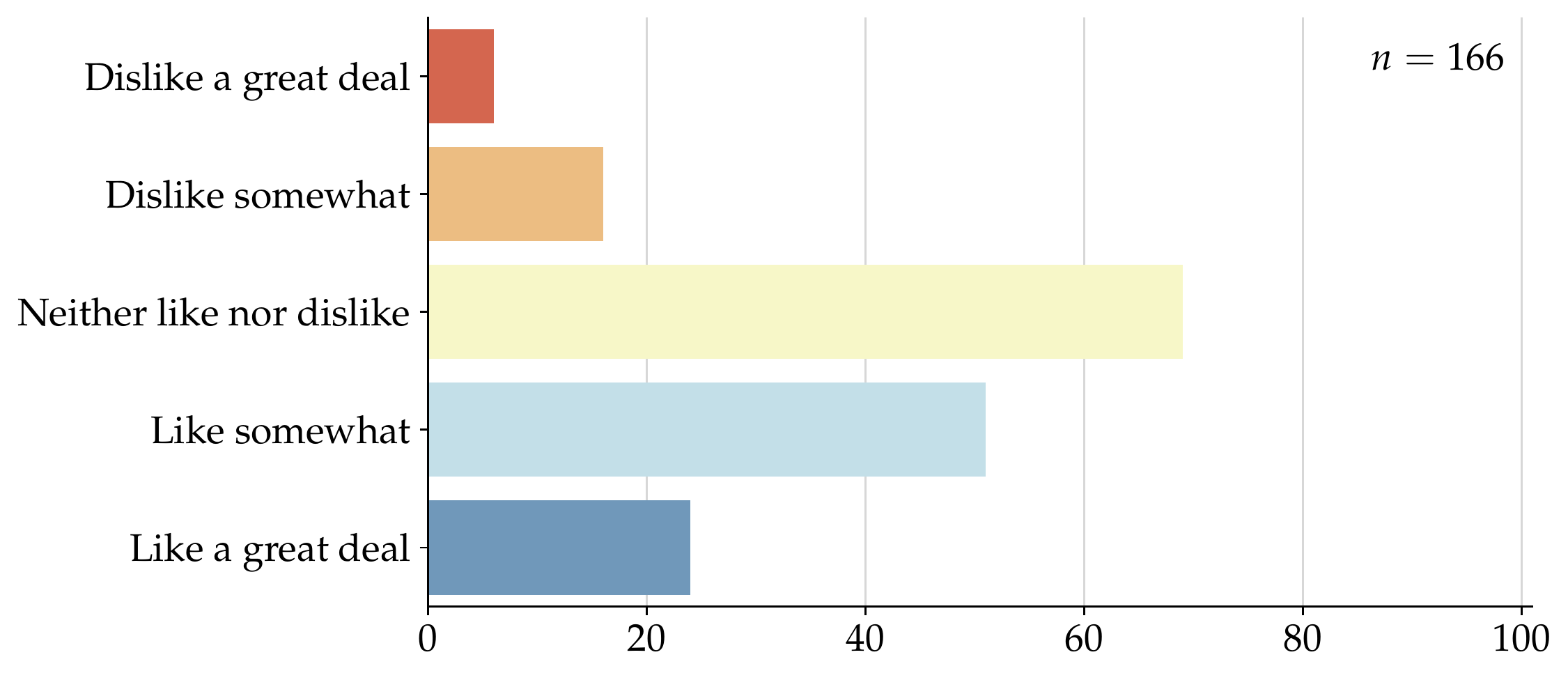}
	
	\question{Why do you not like the ``Extended Abstract'' track?}
  
  We received $n = 19$ text responses. The main issues raised by
  commenters concern a~(perceived) lack of quality of extended
  abstracts, with some respondents citing fears of using such extended
  abstracts as a way to perform ``idea registration'' rather than
  in-depth analyses. Moreover, respondents also stated that reviewing
  such submissions is more complex since the standards for acceptance
  would have to be adjusted accordingly.
	
	\question{Why do you like the ``Extended Abstract'' track?}
	
  We received $n = 28$ text responses. Almost every comment highlights
  the possibility to submit early work or preliminary work and get
  quick feedback by the community. Some respondents also consider this
  track to be advantageous to present non-traditional work, such as
  critique papers or papers that focus on highlighting negative results. 

	\question{Anything else you would like to tell us?}

  We received $n = 24$ responses. Many respondents expressed the wish of
  seeing more instances of \LoG, as well as moving to a hybrid format.
  One respondent specifically requested a track for survey papers,
  while another raised frustrations about the OpenReview platform.
  Finally, one respondent provided helpful insights for further
  improving the review quality, in particular as the conference grows.

	\section{Discussion}

  The overall responses of the community and the general interest
  in a second version of the conference paint a positive picture of the
  first instance of \LoG. Analyzing the experiences in more detail, we
  find that \LoG is a microcosm of issues that are known to plague the
  machine learning community at large. These issues, unsurprisingly, are
  predominantly concerns about aspects of peer review, including the
  ensuing discussion between reviewers and authors. We are excited to
  see that, despite \LoG being a ``grassroots conference'' arising
  \emph{from} the community and \emph{for} the community, respondents
  often rate this conference to have provided them with the ``best
  review experience'' so far. Authors conceded that reviewer standards
  were even slightly higher than at comparable conferences, while also
  citing an overall better experience with the review process.

  These positive experiences contrast with some negative experiences
  of authors. An analysis\footnote{%
    Despite the complexity of the API, we find OpenReview to offer an
    unprecedented level of detail for evaluating a conference.
  } of the discussions shows that there are $n = 29$ ``silent papers,''
  i.e., papers with no in-depth discussion between authors and
  reviewers. While $n = 2$ of these papers were eventually accepted
  because of strong reviews---which, in some sense, obviated the need
  for a discussion---this leaves $n = 27$ papers without an exchange. Of
  these papers, $n = 18$ received no comments from authors, meaning that
  the authors did not comment on the reviews. This could indicate
  a misunderstanding regarding the potential utility of a rebuttal, or it
  could mean that authors did not think that the opinions of reviewers
  could be changed. Believing in the autonomy
  of authors, one could say that the review process worked ``as
  designed'' for these $n= 18$ cases: authors sent in their work,
  authors received feedback, but \emph{chose} not to engage further.
  However, this leaves $n = 9$ papers that were eventually rejected
  without reviewers commenting on a rebuttal provided by authors. These
  are clear \emph{failures} of the review process, since we would at the
  very least expect reviewers to explicitly acknowledge the rebuttal. A brief
  comparison to other conferences shows that the relative numbers of
  such papers are extremely low, indicating that overall engagement of reviewers at \LoG was comparatively high. Nevertheless, we will have to improve our
  processes to avoid such breakdowns in communication.

	\section{Suggestions}

  Given the high quality of the majority of reviews, we will continue
  our vetting procedure and strive to select reviewers with the utmost
  care. We will also retain the rating system of reviewers and
  area chairs, which is a cornerstone of the reviewer awards.  While the
  effect of monetary rewards cannot be fully assessed in our current survey
  setup, we will nevertheless keep this as one feature of \LoG for the
  next instance of the conference. To further focus on review quality,
  we will improve the monitoring of the review process, making use of
  the OpenReview API to find and identify ``silent papers'' early during
  the review process. We will also raise this topic with area chairs so
  that they can better stir and steer such conversations,  ensuring that no discussion
  items are left unanswered.

  One of the insights that we have to tackle on a much broader level
  involves a better tracking of reviewers. While \LoG already uses
  reviewer ratings,\footnote{%
    These ratings are to be taken with a grain of salt, though, since
    the \emph{outcome} of the review phase constitutes a strong
    confounding variable. Authors whose papers are rejected may not be
    willing to concede that they received high-quality reviews.
  }
  it would be beneficial for the whole machine learning community to
  adopt a \emph{reviewer reputation} system. Such a system would
  increase the accountability of reviewers and also serve to highlight
  those that exhibit ``good scientific citizenship.'' Beyond monetary
  awards for a selected set of reviewers, it would be interesting to
  discuss general reviewer compensation. However, instituting such
  a system is a policy change fraught with additional questions (as well as administrative and fiscal complications). 
  While
  it is likely that a proper contract with remuneration would further
  improve review quality, the contract would also need to be enforced if need
  be. 
  This suggests the use of impartial and trusted experts to
  carefully \emph{check} reviews of a conference (raising the follow-up problem of establishing guidelines for identifying, recruiting, remunerating, and overseeing these experts). 
  For \LoG, we will
  ensure that organizers perform this job during the next
  iteration, so that they can engage with problematic reviewers
  or authors early on in the reviewing process.

  \printbibliography
\end{document}